\documentclass[conference]{IEEEtran}
\IEEEoverridecommandlockouts
\usepackage{cite}
\usepackage{amsmath,amssymb,amsfonts}
\usepackage{cancel}
\usepackage{graphicx}
\usepackage{textcomp}
\usepackage{xcolor}
\usepackage{multirow}
\usepackage{amsmath}
\usepackage{tabularray}
\usepackage{booktabs}
\usepackage{makecell}
\usepackage{array}
\usepackage{tabularx}
\usepackage{amsmath}
\usepackage{gensymb}
\usepackage{yhmath}
\usepackage{algpseudocode}
\usepackage{algorithm}
\usepackage[dvipsnames]{xcolor}
\usepackage{url}

\def\BibTeX{{\rm B\kern-.05em{\sc i\kern-.025em b}\kern-.08em
    T\kern-.1667em\lower.7ex\hbox{E}\kern-.125emX}}

\definecolor{antiquefuchsia}{rgb}{0.57, 0.36, 0.51}

\definecolor{darkgreen}{rgb}{0.0, 0.6, 0.0}

\begin{document}

\title{PinPoint: Monocular Needle Pose Estimation for Robotic Suturing via Stein Variational Newton and Geometric Residuals}

\author{Jesse F. d'Almeida\textsuperscript{1},
    Tanner Watts\textsuperscript{2},
    Susheela Sharma Stern\textsuperscript{1},
    James Ferguson\textsuperscript{3}, \\
    Alan Kuntz\textsuperscript{2}, 
    and Robert J. Webster III\textsuperscript{1}
    
    \thanks{
    Research reported in this publication was supported by the Advanced Research Projects Agency for Health (ARPA-H) under the ALISS project, Award Number D24AC00415-00. The ARPA-H award of up to \$11,935,038 provided 50\% of the financial support for this work and National Science Foundation Graduate Research Fellowship Program provided 50\% under Grant No. 1937963 \& 2444112. The opinions and findings in this paper are those of the authors and do not necessarily represent the official views ARPA-H nor the National Science Foundation. (corresponding author email: {jesse.f.dalmeida}@vanderbilt.edu).}
    \thanks{\textsuperscript{1}Department of Mechanical Engineering, Vanderbilt University, TN, USA.}
    \thanks{\textsuperscript{2}Department of Electrical and Computer Engineering and Department of Computer Science, Vanderbilt University, TN, USA.}
    \thanks{\textsuperscript{3}The Robotics Center and the Kahlert School of Computing, University of Utah, Salt Lake City, UT, USA.}
}

\maketitle

\begin{abstract}


Reliable estimation of surgical needle 3D position and orientation is essential for autonomous robotic suturing, yet existing methods operate almost exclusively under stereoscopic vision. In monocular endoscopic settings, common in transendoscopic and intraluminal procedures, depth ambiguity and rotational symmetry render needle pose estimation inherently ill-posed, producing a multimodal distribution over feasible configurations, rather than a single, well-grounded estimate. We present \textbf{PinPoint}, a probabilistic variational inference framework that treats this ambiguity directly, maintaining a distribution of pose hypotheses rather than suppressing it. PinPoint combines monocular image observations with robot-grasp constraints through analytical geometric residuals with closed-form Jacobians. This framework enables efficient Gauss-Newton preconditioning in a Stein Variational Newton inference, where second-order particle transport deterministically moves particles toward high-probability regions while kernel-based repulsion preserves diversity in the multimodal structure. 
On real needle-tracking sequences, PinPoint reduces mean translational error by 80\% (down to 1.00\,mm) and rotational error by 78\% (down to 13.80\,$\degree$) relative to a particle-filter baseline, with substantially better-calibrated uncertainty. 
On induced-rotation sequences, where monocular ambiguity is most severe, PinPoint maintains a bimodal posterior 84\% of the time, almost three times the rate of the particle filter baseline, correctly preserving the alternative hypothesis rather than committing prematurely to one mode. Suturing experiments in ex vivo tissue demonstrate stable tracking through intermittent occlusion, with average errors during occlusion of 1.34\,mm in translation and 19.18\,$\degree$ in rotation, even when the needle is fully embedded.

\end{abstract}

\section{Introduction}
Suturing is one of the most fundamental and cognitively demanding tasks in Robot-assisted Minimally Invasive Surgery (RMIS)\cite{hubens_performance_2003}. The task demands repeated, precise manipulation of a curved needle under constrained instrument motion \cite{faraz2000engineering}; even expert surgeons face significant cognitive and physical burden, contributing to procedural variability and operative fatigue \cite{sen_automating_2016, pedram_autonomous_2021}.
Automating components of this process could reduce that burden and improve procedural consistency, motivating substantial research into suturing automation \cite{wilcox_learning_2022}.


\begin{figure}
    \centering
    \includegraphics[width=\columnwidth]{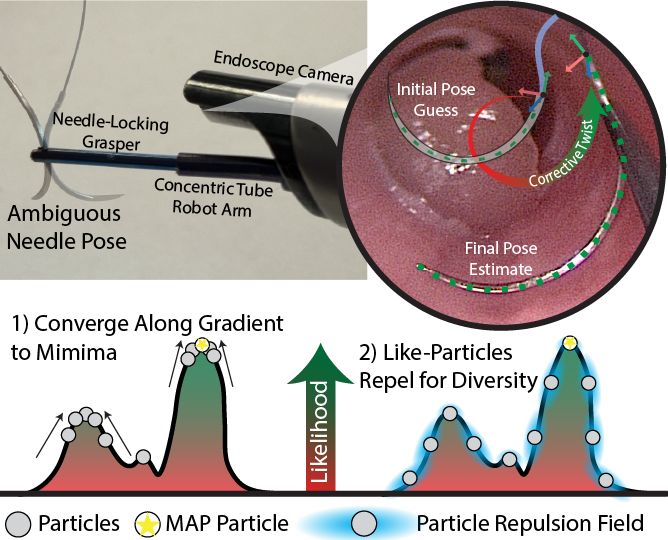}
    \caption{Conceptual illustration of the proposed monocular needle tracking framework. (Top) Image and robot-based observations define an error over the needle pose estimate, which is iteratively solved for with PinPoint. The isometric robot view highlights the inherent depth ambiguity of monocular needle pose estimation, where multiple 3D configurations can produce similar image projections. (Bottom) PinPoint uses Stein Variational Inference, which represents the posterior over needle poses using a set of interacting particles. Particles are deterministically transported toward high-probability regions of the posterior while repulsive interactions prevent collapse and preserve multimodal hypotheses. The maximum a posteriori (MAP) particle corresponds to the most likely needle pose estimate.}
    \label{fig:intro}
\end{figure}

Most autonomous suturing research has targeted laparoscopic environments with stereoscopic visualization and rigid instrumentation \cite{shkurti_systematic_2025}. Stereoscopic depth information makes the pose estimation problem tractable: they constrain the 3D solution well enough that probabilistic filters can recover a single reliable pose estimate \cite{chiu_markerless_2022, ozguner_visually_2021}. Many clinically important procedures, however, require compact instrumentation through narrow anatomical pathways where only monocular visualization is available, and no such depth cues exist \cite{hendrick_hand-held_2015, harvey_novel_2020}. Concentric tube robots (CTRs) are one platform well-suited to these settings \cite{webster_iii_design_2010, dupont_design_2010}, and have been demonstrated in transendoscopic and intraluminal procedures precisely where monocular perception constraints apply \cite{hendrick_hand-held_2015, dalmeida_prostatectomy_2026}.

The central challenge of monocular needle pose estimation is not merely that depth is unobserved, but the presence of multiple geometrically consistent solutions (multimodal) and often weak observability along certain directions (ill-conditioning).
Projection ambiguity, partial occlusions, and the rotational symmetry of curved needles can produce multiple 3D configurations consistent with the same image, inducing a multimodal distribution over feasible poses that no single deterministic estimate can represent \cite{manhardt2019explaining, wehrbein2021probabilistic}. 
In stereoscopic systems, depth cues resolve this ambiguity and enable probabilistic filters to recover a single reliable estimate \cite{chiu_markerless_2022, chiu_real-time_2022, ozguner_visually_2021}; in the monocular setting, the ambiguity remains.
Existing monocular approaches address this by reprojection or with learning-based detection \cite{iyer_single_2013, li_monocular_2024}. Neither incorporate robot kinematics or grasp constraints, leaving the estimate grounded solely in the image observations and unable to represent the multimodal structure of the posterior.

This paper presents \textbf{PinPoint}, a probabilistic variational inference framework for real-time monocular needle pose estimation. PinPoint formulates needle tracking as inference over $SE(3)$, maintaining a distribution of pose hypotheses that preserves the genuine multimodal structure of the posterior rather than committing to a single estimate.
Geometric constraints from monocular image observations and the physical robot-grasp are expressed as analytical residuals with closed-form Jacobians, maintaining the pose estimation grounded in 3D even when depth is unobserved. These residuals drive a Stein Variational Newton (SVN) inference \cite{detommaso_stein_2018}, which uses Gauss-Newton curvature to precondition a deterministic particle transport, enabling efficient convergence while kernel-based repulsion preserves hypothesis diversity across modes as well as an estimate for its uncertainty.

PinPoint is evaluated on real needle-tracking sequences with ground-truth poses obtained from a rigidly mounted checkerboard. Across sequences of controlled motion, induced out-of-plane rotations, and ex vivo suturing with severe occlusions, PinPoint substantially outperforms a particle filter baseline in accuracy, uncertainty calibration, and multimodal posterior representation.

\section{Related Works}

\subsection{Stereo-Based Needle Pose Estimation}

Stereo-based needle pose estimation exploits the elliptical reprojection of a circular needle to recover 3D pose from image correspondences, forming the backbone of influential autonomous systems \cite{sen_automating_2016, pedram_autonomous_2021}. 
Subsequent work introduced probabilistic filtering frameworks that fuse multiple geometric observation models with robot motion priors improve robustness under noise and partial occlusions \cite{chiu_markerless_2022, ozguner_visually_2021, ozguner_three-dimensional_2018}, as well as grasp-awareness that reject infeasible poses by encoding physical needle-gripper constraints \cite{chiu_real-time_2022}. Despite these advances, all such methods rely on stereoscopic depth cues, limiting their applicability in monocular endoscopic settings. PinPoint builds on the geometric observation models developed in this line of work to the monocular case, substituting stereo depth with robot-grasp constraints that provide 3D grounding from a single view. 

\subsection{Monocular Needle Pose Estimation}

Monocular pose estimation presents a more constrained version of the pose estimation problem: without depth cues, purely geometric approaches that reproject a circular needle model onto the image plane are underconstrained and sensitive to noise \cite{iyer_single_2013}.

Recent learning-based methods have made progress on this front, though important gaps remain. 
Li et al. \cite{li_monocular_2024} avoid the circular shape assumption entirely, using a multi-task network to extract segmentation masks and tip/tail keypoints from which 6-DoF pose is recovered via point-to-mask distance minimization. 
The method is designed for generality by excluding robot kinematics and grasp constraints, it targets scenes where the needle may not be grasped.
Wang et al. \cite{wang_monocular_2025} take a more direct approach, training a CNN on a hybrid real/synthetic dataset to predict three needle keypoints from which 3D orientation is recovered using the needle's known geometry, though 3D position remains unaddressed.

Neither approach incorporates robot kinematics or grasp constraints, leaving the pose estimate grounded solely in image observations. More fundamentally, neither represents the multimodal posterior structure that monocular depth ambiguity induces, rather, both produce a single deterministic estimate with no mechanism for understanding uncertainty or maintaining alternative hypotheses. 
PinPoint addressed both gaps: its uses a learned detector only to extract image observations, with all pose estimation performed through geometric residuals grounded in physical measurement models; and it explicitly maintains a distribution over pose hypotheses, preserving the multimodal structure rather than collapsing it.

\subsection{Probabilistic and Multimodal Pose Inference}

The circular arc geometry of a suture needle admits multiple geometrically consistent 3D configurations from a single monocular view as a consequence of the well-known two-solution ambiguity of a circular pose estimation when projected onto the image plane \cite{zheng2008another}. Thus, the pose posterior becomes multimodal. Prior work in monocular surgical tool pose estimation has obsevred the practical consequences of this structure wherein ambiguous configurations produce high errors \cite{barragan2024realistic}. Methods that commit to a single estimate mode cannot represent this structure by construction, and analogous failure modes have been documents in camera relocalization \cite{bui20206} and monocular human pose estimation \cite{wehrbein2021probabilistic}, whereby maintaining full posterior distributions substantially outperformed point estimation approaches. PinPoint addresses this directly by treating needle tracking as multimodal inference over $SE(3)$, rather than patching a point estimator with heuristic priors.

Particle filters have been the dominant tool for probabilistic needle
tracking \cite{chiu_markerless_2022, ozguner_visually_2021}, but suffer from importance weight collapse under tight geometric constraints. When the likelihood is sharply peaked, most particles receive negligible weight, leading to degeneracy and loss of hypothesis diversity \cite{liu_stein_2016}. Stein Variational Gradient Descent (SVGD) \cite{liu_stein_2016} addresses this by deterministically transporting particles toward high-probability regions while kernel-based repulsion maintains diversity. Stein Variational Newton (SVN) \cite{detommaso_stein_2018} further incorporates Hessian preconditioning to adapt transport to local posterior curvature, enabling rapid convergence even in ill-conditioned settings. PinPoint adopts SVN as its inference method, deriving closed-form Gauss-Newton Hessian approximations from analytical residual Jacobians to ground particle transport directly in the geometry of the needle pose problem.

\section{Formulation and Conditional Likelihoods}
\label{sec: problem}

\begin{figure}
    \centering
    \includegraphics[width=\columnwidth]{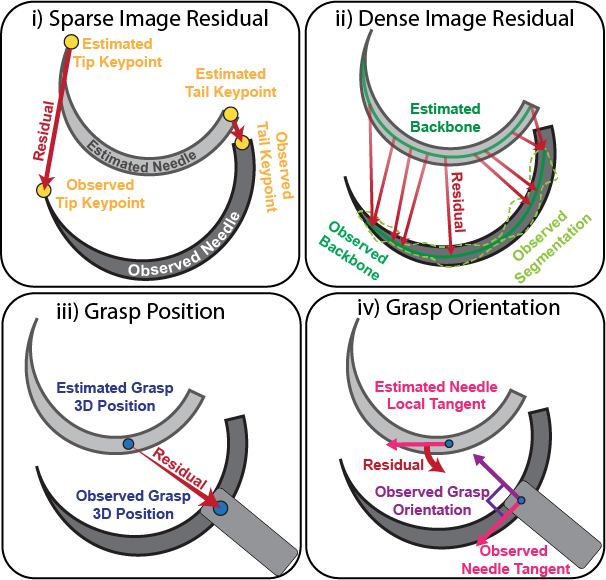}
    \caption{Sources of information used for monocular needle pose estimation and the geometric reach of their corresponding residuals: (i) image keypoints at the needle tip and tail, (ii) needle backbone segmentation, (iii) the C-Lock rigidly fixes the needle at a known grasp point, and (iv) the C-Lock enforces a perpendicular relationship between the gripper axis and the needle tangent at the grasp location.}
    \label{fig: residuals}
\end{figure}

We consider the problem of estimating the full 6 degree-of-freedom (DoF) needle pose $T \in \mathrm{SE}(3)$, which maps needle frame coordinates to the optical camera frame. 
PinPoint draws on two complementary sources of information: monocular image observations (top row of Fig. \ref{fig: residuals}) and robot grasper kinematic constraints (bottom row of Fig. \ref{fig: residuals}).

From the current image, we extract two types of observations. 
First, learned sparse keypoints (2D detections of the needle tip and tail) provide known correspondences to points on the canonical needle model, yielding strong localized constraints on image-plane position and orientation. 
These are complemented by a dense needle backbone segmentation, which provides global shape constraints and remains informative even under partial occlusion.
Finally, PinPoint incorporates uncertain robot grasper kinematics from the needle grasp, which provide complementary geometric constraints and help anchor the depth of the needle pose, which is typically ill-conditioned under monocular observations alone.

Together, these measurements define the following overall Negative Log-Likelihood (NLL) objective for needle pose inference:
\begin{equation}
\label{eq: least squares}
\mathcal{L}(T) = \mathcal{L}_{\mathrm{sparse}}(T) + \mathcal{L}_{\mathrm{dense}}(T) + \mathcal{L}_{\mathrm{grasp}}(T),
\end{equation}
where $\mathcal{L}_{\mathrm{sparse}}(T)$ corresponds to sparse keypoint reprojection error, $\mathcal{L}_{\mathrm{dense}}(T)$ captures alignment with the segmented needle backbone, and $\mathcal{L}_{\mathrm{grasp}}(T)$ enforces consistency with the (uncertain) robot kinematics.
Each likelihood cost is constructed from a residual $r_k(T) = h_k(T) - z_k$ as $\mathcal{L}_k(T) = r_k^\top \Sigma_k^{-1}r_k$, where $h_k$ is the measurement model and $z_k$ is the observation. 
The next sections describe how each likelihood is specifically constructed, how the system is linearized, and how we efficiently approximate a potentially multimodal posterior distribution over needle poses.







\subsection{Sparse Keypoint Likelihoods}

When detectable, sparse visual features provide strong localized constraints on the needle pose. 
Building on the \textit{Points Matching to Ellipse} model of Chiu et al. \cite{chiu_markerless_2022}, we use detected needle endpoints as direct reprojection constraints.
Specifically, we may detect pixel keypoints corresponding to the needle tip and tail, denoted $z_{\mathrm{tip}}, z_{\mathrm{tail}} \in \mathbb{R}^2$, which have known correspondences to the 3D points $p_{\mathrm{tip}}, p_{\mathrm{tail}} \in \mathbb{R}^3$ in the needle frame.

\begin{figure}
    \centering
    \includegraphics[width=.8\columnwidth]{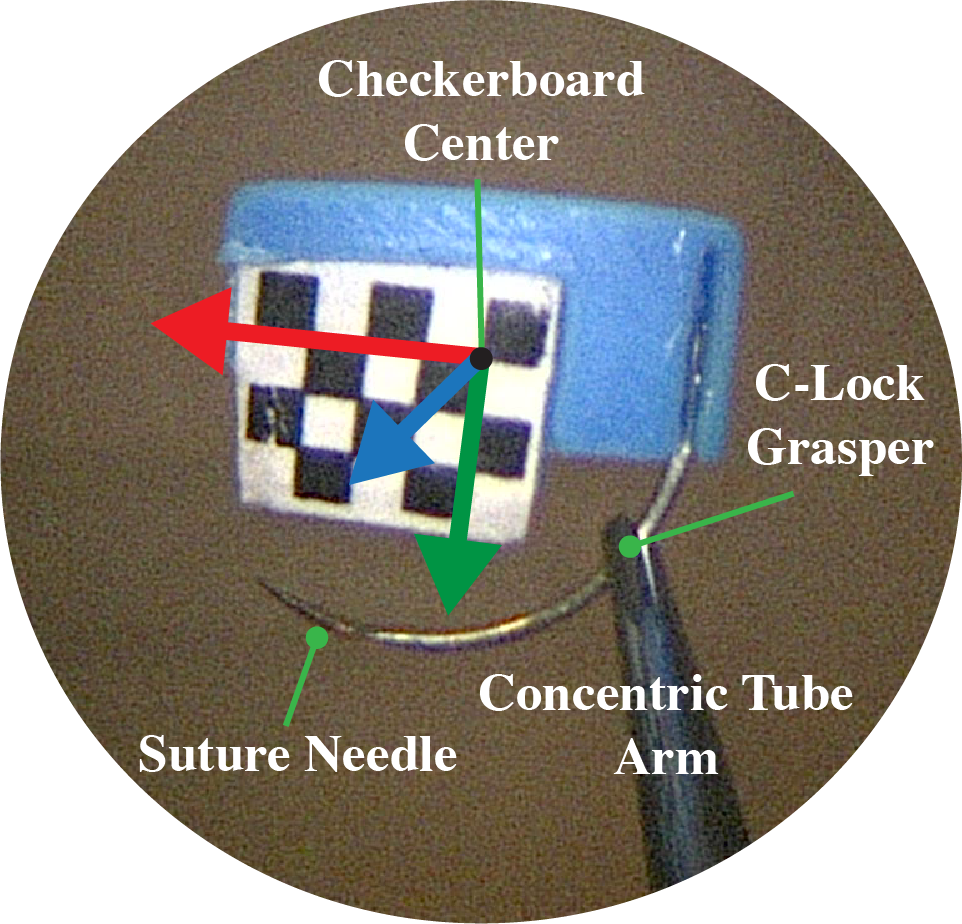}
    \caption{The needle is rigidly fixed to the checkerboard such that the transform between the centers can be measured, enabling ground truth needle pose measurements, for calibration and accuracy tests.}
    \label{fig:checkerboard}
\end{figure}

Given the needle pose $T$, the pixel location of a homogeneous 3D point $X = \begin{bmatrix} p, 1\end{bmatrix}^T$ is $\pi(TX)$, where $\pi(p)=[f_x x/z + c_x, f_y y/z + c_y]^T$, $(f_x, f_y)$ are the focal lengths, $(c_x, c_y)$ is the camera center, and $p=[x, y, z]^T$ is the point in optical frame coordinates. 
We assume that the keypoint measurement is corrupted by Gaussian noise so that the NLL over all detected keypoints is:
\begin{equation}
\begin{gathered}
    \mathcal{L}_{\mathrm{sparse}}(T) = \sum_i \| r_{\mathrm{key}, i} \|_{\Sigma_{\mathrm{key}}}
    ,\\
    \quad
    r_{\mathrm{key}, i} = \pi(TX) - z_{\mathrm{key}, i}
    .
\end{gathered}
\end{equation}
where $ \|\cdot \|_\Sigma$ denotes the Mahalanobis distance with respect to the covariance. Note that all noise covariances, including that of the keypoints $\Sigma_{\mathrm{key}}$, are calibrated empirically by evaluating each residual at the ground-truth needle pose, as shown in Fig. \ref{fig:checkerboard}, across a calibration dataset, as described in Section \ref{sec: noise calibration}.
Additionally, we show how to linearize this likelihood and all other measurement models in Appendix \ref{sec: jacobian_definitions}.


\subsection{Dense Segmented Backbone Likelihood} 

When keypoints are occluded or undetectable, our dense backbone segmentation supplies information without requiring point-to-point correspondence.
We model the needle as a 3D circular arc of known radius $r$; its projection is dependent on the needle pose $T$, which induces the image conic $C(T)$.
Overall, the projected conic $C(T)$ overlapping with the segmentation corresponds with high likelihood.

The image conic $C(T)$ follows from classical quadric projection theory \cite{cross_quadric_1998}, where the the planar quadric $Q_{\mathrm{plane}} = \mathrm{diag}(1,1,-r^2)$ is transformed by the projective homography:
\begin{equation}
\label{eq: H def}
    H(T) = K\,[I~|~0]\,T\,M \in \mathbb{R}^{3\times3},
    \quad 
    M = \begin{bmatrix}
            1 & 0 & 0\\
            0 & 1 & 0\\
            0 & 0 & 0\\
            0 & 0 & 1\\
        \end{bmatrix}
\end{equation}
where $K$ is the camera intrinsic matrix and $M$ embeds the planar circle into homogeneous 3D coordinates, yielding:
\begin{equation}
     C(T) = H(T)^{-T} Q_{plane} H(T)^{-1}
     .
\end{equation}

Given homogeneous image points $x_j = [u_j, v_j, 1]^T$ from the segmented backbone, a natural way to quantify consistency between the conic $C(T)$ and $x_j$ is:
\begin{equation}
    f_i(T) = x_j^T C(T) x_j
    ,
\end{equation}
which is zero for points lying exactly on the conic. 
However, this algebraic distance depends on the arbitrary scale of the conic parameters and does not correspond directly to geometric image distance.
To address this, we instead use the Sampson distance, a first-order approximation of the geometric reprojection error \cite{hartley2003multiple}:
\begin{equation}
\begin{gathered}
r_{\mathrm{dense}, j}(T) =
\frac{x_j^\top C(T) x_j}
{\left\| \nabla_x \big(x_j^\top C(T) x_j\big) \right\|_2}
,\\
\mathrm{where} \quad
\nabla_x \big(x^\top C x\big) = (C + C^\top)x
.
\end{gathered}
\end{equation}
Here, the normalization converts the algebraic conic constraint into an approximate Euclidean image distance, preserving differentiability for Gauss-Newton updates.

We model each image-point segmentation measurement with Gaussian noise $e_{d,i} \sim \mathcal{N}(0,\sigma_d^2)$, yielding the following for the dense segmentation NLL cost:
\begin{equation}
\begin{gathered}
    \mathcal{L}_{\mathrm{dense}}(T) = \sum_j \| r_{\mathrm{dense}, j} \|_{\Sigma_\mathrm{dense}}
    .
\end{gathered}
\end{equation}
Again, the covariance $\Sigma_\mathrm{dense}$ is calibrated in Section \ref{sec: noise calibration}, and linearization is carried out in Appendix \ref{sec: jacobian_definitions}.


\subsection{Robot Kinematics Likelihood}

In contrast to image-based observations, the robot kinematics provides direct 3D information that remains available even under visual occlusion.
Specifically, the C-Lock \cite{dalmeida_stent_2026} is a grasper for CTRs that rigidly fixes the needle at a known point relative to the robot.
This combined with the robot kinematics provides the 3D position of the grasp point in the camera frame, directly anchoring the depth of the needle pose. 
The grasper geometry additionally provides an orientation constraint, as the C-Lock enforces approximate perpendicularity between the gripper axis and the needle tangent at the grasp location.
The C-Lock grasper rigidly holds the needle at a known point on the needle, denoted $X_{\mathrm{grasp}}$. 
Unlike traditional forceps graspers, where the needle may contact the jaws anywhere along the surface manifold, requiring an additional solve for the grasp point \cite{chiu_real-time_2022}, the C-Lock fixes the needle at a known pose relative to the grasper with no additional unknowns introduced.

Although the specific point $X_{\mathrm{grasp}}$ on the canonical needle may vary between grasps, the transform between the grasper frame and the grasp point remains rigid while the needle is held during use.
Forward kinematics provides an estimate $p_{\mathrm{grasp}}$ of the grasp position relative to the camera, which we can compare with the needle pose $T$ to get the position residual:
\begin{equation}
    r_p(T) =  T X_{\mathrm{grasp}} - p_{\mathrm{grasp}}
    .
\end{equation}
%


The grasper geometry further constrains needle orientation, eliminating one rotational DoF within the grasp. 
Specifically, the C-Lock enforces approximate perpendicularity between the grasper axis $\hat{z}_{\mathrm{grasper}}$ and the unit tangent direction $\hat{t}_{\mathrm{grasp}}(T)$ of the needle at $X_{\mathrm{grasp}}$.
This orthogonality constraint is encoded via the following residual:
\begin{gather}
    r_\perp(T) = \hat{z}_{\mathrm{grasper}}^\top \hat{t}_{\mathrm{grasp}}(T)
    .
\end{gather}
This constraint reduces rotational ambiguity and, critically, remains valid during partial and complete visual occlusion, providing orientation information precisely when image observations are unavailable.


These residuals are combined to form the overall robot kinematics grasp NLL:
\begin{equation}
    \mathcal{L}_\mathrm{grasp}(T) = \| r_p \|_{\Sigma_p} + \| r_\perp \|_{\Sigma_\perp}
    .
\end{equation}
Thus, given all of the individual likelihoods in (\ref{eq: least squares}), we can proceed with estimating the conditional distribution over needle poses $T$.

\subsection{Second-Order Conditioning}
\label{sec:svn_inference}

The Gauss-Newton Hessian approximation $H = \sum_k J_k^T J_k$ aggregates curvature information across all four residual types. 
In practice, the conditioning of $H$ varies substantially based on needle pose and observation availability. 
To motivate second-order preconditioning in the inference step (i.e., using the SVN method as opposed to SVGD), we examine $H$ on representative evaluation frames. The pose ambiguity from monocular observations results in an ill-conditioned Hessian, meaning that gradient-only transport, which implicitly treats all 6 pose dimensions equally, takes steps that are too large along well-constrained directions and too small along poorly constrained ones. The Gauss-Newton preconditioning directly addresses this by adapting the step to the shape of the local curvature. We demonstrate this directly by removing the second-order information (effectively an SVGD method) as part of the ablation studies \ref{sec: ablation}.

\section{PinPoint Variational Inference}

\begin{algorithm}
\caption{PinPoint}
\label{alg:svn}
\begin{algorithmic}[1]
\Require Observations $\mathcal{Z}$,
         particles $\{T^{(i)}_0\}_{i=1}^N \subset SE(3)$
\Ensure $\hat{T}$, $\Sigma$

\For{$k = 1$ to $K$}
    \For{$i = 1$ to $N$}
        \State $g^{(i)} \gets -\nabla_{\xi}\mathcal{L}(T^{(i)}_{k-1})$
        \State $H^{(i)} \gets (J^{(i)})^{\top} J^{(i)}$
        \State $\tilde{H}^{(i)}_\epsilon \gets$ eigenvalue-floored $H^{(i)}$
    \EndFor
    \State $h \gets \max\!\left(
        \frac{\mathrm{med}_{i<j}\|\xi^{(i)}-\xi^{(j)}\|^2}{\log N},\,
        \sigma_Q^2\right)$
    \For{$i = 1$ to $N$}
        \State $\Delta \xi^{(i)} \gets \frac{1}{N}\sum_{j}\bigl[
            k(\xi^{(j)}\!,\xi^{(i)})\,
            \tilde{H}^{(j)-1}_\epsilon g^{(j)}
            + \nabla_{\xi^{(j)}} k\bigr]$
        \State $T^{(i)}_{k} \gets \exp(\widehat{\Delta \xi^{(i)}}) T^{(i)}_{k-1}$
    \EndFor
\EndFor
\State $\hat{T} \gets \operatorname*{arg\,max}_{i}\log p(T^{(i)}_K \mid \mathcal{Z})$
\State $\Sigma \gets \mathrm{Cov}\bigl(\{\log(\hat{T}^{-1}T^{(i)}_K)\}\bigr)$
\State \Return $\hat{T}$, $\Sigma$
\end{algorithmic}
\end{algorithm}


While a single maximum-likelihood estimate can be obtained by minimizing the overall NLL $\mathcal{L}(T)$, monocular needle pose estimation is inherently ambiguous under limited depth cues and partial occlusion, and the resulting posterior may be multimodal. 
Therefore, to represent general distributions, PinPoint uses a Stein Variational Newton (SVN) framework, which approximates the posterior using a set of interacting particles that are deterministically transported toward high-probability regions \cite{detommaso_stein_2018}.
Unlike particle filters, SVN leverages analytic gradients together with curvature information (e.g., Gauss--Newton approximations to the Hessian) to move particles coherently, while kernel-based repulsion preserves diversity and enables the representation of multiple modes. 
In PinPoint, the Gauss--Newton curvature is used to precondition the Stein updates, yielding efficient second-order inference grounded in the underlying problem geometry.
This formulation enables SVN to capture multimodal posteriors with relatively few particles compared to sampling-based approaches, while maintaining consistency with the underlying measurement models.


\subsection{Stein Variational Update}

We approximate the posterior $p(T \mid \mathcal{Z})$ using a set of $N$ particles $\{T_i\}^N_{i=1} \subset SE(3)$; Stein variational inference deterministically transports this particle set to approximate the posterior by minimizing the Kullback–Leibler (KL) divergence between the empirical particle distribution and $p(T \mid \mathcal{Z})$.
Each particle is locally updated via a left-multiplicative perturbation in the tangent space of $SE(3)$. Specifically, at each iteration a twist update $\Delta \xi_i \in \mathbb{R}^6$ is applied as:
\begin{equation}
    T_i \leftarrow \exp(\widehat{\Delta \xi_i})\,T_i
    ,
\end{equation}
All Stein updates are performed in this tangent space and then mapped back to $SE(3)$ through the exponential map following standard Lie group optimization conventions \cite{barfoot_associating_2014}.

At each iteration, the particles are updated to minimize KL divergence by balancing attraction toward regions of high posterior probability with repulsive interactions that preserve particle diversity. 
We first compute the standard Stein variational velocity in the tangent space:
%
%
\begin{equation}
\begin{split}
    \phi(\xi_i) = \frac{1}{N}\sum_{j=1}^N
    \Bigl[
    &\underbrace{k(\xi_j,\xi_i)\,\nabla_{\xi_j}\log p(\xi_j\mid\mathcal{Z})}_{\text{attraction}}\\
    &+\underbrace{\nabla_{\xi_j}k(\xi_j,\xi_i)}_{\text{repulsion}}
    \Bigr]
\end{split}
\label{eq:stein_velocity}
\end{equation}
where $k(\cdot, \cdot)$ is a positive-definite kernel that governs inter-particle repulsion.
Evaluation of the log-posterior and its gradient is defined in Section \ref{sec: problem}.


\subsection{Whitened Kernel}
Inter-particle distances are computed in a whitened tangent space to account  for the scale disparity between translational and rotational degrees of  freedom. Let $S = \mathrm{diag}(\sigma_t \mathbf{1}_3, \sigma_r \mathbf{1}_3)$  be a diagonal scaling matrix with translational scale $\sigma_t$ (meters)  and rotational scale $\sigma_r$ (radians). The kernel is:
\begin{equation}
    k(\xi_j, \xi_i) = \exp\!\left(
        -\frac{\|S^{-1}(\xi_i - \xi_j)\|^2}{2h}
    \right)
\end{equation}
The bandwidth $h > 0$ is set via the median heuristic: $h = m / \log N$,  where $m$ is the median pairwise squared distance in the whitened tangent  space. To prevent bandwidth collapse when particles cluster tightly, which  would suppress repulsion precisely when diversity is most needed, $h$ is  floored at the mean process noise variance:
\begin{equation}
    h = \max \!\left(\frac{m}{\log N}, \; \sigma_Q^2\right),
    \qquad
    \sigma_Q^2 = \frac{1}{6}\,\mathrm{tr}(Q),
\end{equation}
where $Q$ is the process noise covariance. When $S = I$ with isotropic median-heuristic bandwidth, this reduces to the standard SVGD kernel \cite{liu_stein_2016}, which serves as an ablation baseline in Section \ref{sec:ablation}.


\subsection{Gauss-Newton Preconditioning}

To improve convergence and account for local curvature of the problem, we precondition the Stein update using second-order information derived from a Gauss–Newton approximation of the Hessian following \cite{detommaso_stein_2018}. 
For each particle $i$, we form:
\begin{equation}
    \label{eq:svn_preconditioned_update}
    H_i \approx J_i^\top J_i,
    \qquad
    \Delta \xi_i = H_i^{-1}\,\phi(\xi_i),
\end{equation}
where $J_i$ is the stacked Jacobian of all residuals evaluated at particle $i$. This yields a Newton-like transport toward high-probability regions of the posterior on a per-particle basis, incorporating second-order geometric information at negligible additional cost since the Jacobians are already required for gradient computation. 

The Gauss-Newton Hessian $H_i = J_i^\top J_i$ is positive semi-definite by construction, and in monocular needle pose estimation. Additionally, it is structurally rank-deficient due to the unobservable directions, specifically the depth along the optical axis and the out-of-plane rotations, which produce near-zero eigenvalues regardless of observation quality. Scalar damping $H_i + \lambda I$, as is common when inverting, inflates all eigenvalues uniformly and does not respect this geometric structure. 

Instead, we solve via truncated eigen decomposition of the symmetrized Hessian $\tilde{H}_i = \frac{1}{2}(H_i + H_i^\top)$ \cite{nocedal2006numerical}. Let $\tilde{H}_i = V_i\,\mathrm{diag}(w_i)\,V_i^\top$ be its decomposition. We regularize by flooring each each eigenvalue relative to the largest:
\begin{gather}
    \Delta \xi_i = V_i\,\mathrm{diag}(\tilde{w}_i)^{-1}\,V_i^\top\,\phi(\xi_i),
    \\
    \tilde{w}_{i,k} = \max\!\left(w_{i,k},\;\epsilon\cdot w_{i,\mathrm{max}}\right),
\end{gather}
where $\epsilon = 10^{-2}$. This bounds the effective condition number at  $\epsilon^{-1} = 100$, suppressing steps along unobservable directions without eliminating them. 




\subsection{Pose Extraction}
After convergence of the Stein updates, the particle set provides a nonparametric approximation of the posterior distribution over $SE(3)$. The complete estimation procedure is summarized in Algorithm \ref{alg:svn}. 
When a single pose estimate $\hat{T}$ is required rather than a distribution, we use the maximum a posteriori (MAP) particle with the highest likelihood after iteration.



\subsection{Uncertainty Estimation}
Given the converged particle set $\{T^{(i)}\}_{i=1}^N$, we quantify pose uncertainty directly from the posterior approximation. 
Using the MAP estimate $\hat{T}$ as a reference pose, we compute tangent-space errors:
\begin{equation}
    \xi^{(i)} = \log\!\left(T_{\mathrm{ref}}^{-1}T^{(i)}\right)\in\mathbb{R}^6,
\end{equation}

which are expressed as translational and rotational components, where $\xi_{1:3}$ denotes translation and $\xi_{4:6}$ denotes the axis-angle rotation vector. Translational and rotational magnitudes are summarized by:
\begin{equation}
r_t^{(i)}=\|\xi_{1:3}^{(i)}\|_2,\qquad r_r^{(i)}=\|\xi_{4:6}^{(i)}\|_2,
\end{equation}

We report the weighted 95th percentile of the particle distribution as scalar uncertainty measures for translation (m) and rotation (rad), respectively:
\begin{align}
    q_{95}^t = \mathrm{qQuantile}(\{r_t^{(i)}\}_{i=1}^N, 0.95),\\
    q_{95}^r = \mathrm{qQuantile}(\{r_r^{(i)}\}_{i=1}^N, 0.95),
\end{align}

We additionally compute a local tangent-space covariance as a second-order summary of the particle distribution:
\begin{equation}
    \Sigma =
\sum_{i=1}^N w_i \,
(\xi^{(i)} - \bar{\xi})
(\xi^{(i)} - \bar{\xi})^T,
\qquad
\bar{\xi} = \sum_{i=1}^N w_i \xi^{(i)}.
\end{equation}

\section{Experiments}
We evaluate PinPoint on monocular needle tracking tasks representative of robotic suturing scenarios. Experiments are performed on the Virtuoso Endoscopy System (VES) (Virtuoso Surgical, Nashville, TN, USA)~\cite{virtuoso-ves-manual}, a dual-arm concentric tube robot. The right arm is equipped with the C-Lock \cite{dalmeida_stent_2026} and is used for all needle manipulation.

The needle geometry is known a priori and fixed across all experiments. We used a 4-0 Reverse Cutting suture (P-3 Polyamide 13\,mm 3/8c, Ethicon Inc, Raritan, NJ, USA), which has cross-sectional diameter of 0.4\,mm and a radius of curvature of 4.5\,mm. 

All experiments are conducted using a calibrated monocular endoscopic camera observing a rigidly grasped needle. The camera provides monocular RGB images at a fixed resolution ($1080\times1080$ px) and frame rate ($30$~Hz). Image observations consist of sparse keypoint detections corresponding to the needle tip and tail, as well as dense segmentations of the visible needle backbone. The keypoints are determined using a small U-Net model while the segmentation model used was developed in \cite{isik2026benchmark}, and runs at a frequency of $10$Hz.

Prior works primarily evaluate monocular needle pose estimation in simulation through synthetic noise injection \cite{chiu_markerless_2022, ozguner_three-dimensional_2018}. In contrast, we obtain real-world ground truth pose measurements using a physical calibration target. Specifically, we designed a mount that rigidly attaches a suture needle to a $3\times4$ checkerboard with $1$~mm squares, while leaving the majority of the needle length free for grasping and manipulation. This setup is shown in Fig.~\ref{fig:checkerboard}. By detecting the checkerboard using OpenCV \cite{opencv_library} and measuring the fixed transform between the checkerboard center and the needle coordinate frame, we obtain ground truth needle poses at each image frame. Note that the checkerboard is used only for accuracy evaluation and is not a part of the PinPoint framework; we segment the checkerboard out of the mask entirely. 

\subsection{Experimental Datasets}

Using the VES platform and the ground-truth needle frame, four experimental conditions were evaluated, each targeting a distinct aspect of tracking performance. 

\noindent\textbf{Slow-speed tracking:} The slow-speed dataset was collected at approximately 1\,mm/s (half of normal-speed), with motion restricted to one axis of motion at a time (left/right, up/down, in/out, and tool rotation) to isolate individual DoFs. This condition provides a controlled baseline for evaluating accuracy and uncertainty calibration under favorable visual conditions.

\noindent\textbf{Normal-speed tracking:} The normal-speed dataset reflects the pace as which an experienced used operates during suturing, approximately 2\,mm/s. This condition provides a controlled baseline for evaluating accuracy and uncertainty calibration under realistic movement dynamics, where the motion was not constrained. 

\noindent\textbf{Induced-rotation sequences:} A separate set of sequences was collected in which the needle undergoes induced, large out-of-plane rotations by external forces. These induced rotations break one of the geometric constraints of the C-Lock, which keeps the needle rotationally aligned unless in the presence of outside forces. A circular needle viewed monocularly can admit multiple geometrically consistent 3D poses that produce nearly identical image projections, and this ambiguity intensifies as the needle rotates away from the image plane. These sequences specifically target the ability of each method to represent multimodal posteriors rather than committing prematurely to a single hypothesis. 


\noindent\textbf{Suturing in ex vivo tissue:} Finally, PinPoint is evaluated during a suturing task in which a curved needle is driven through ex vivo chicken breast tissue. This condition introduces realistic challenges including partial occlusions, rapid needle motion, and changes in visible needle geometry during tissue interaction. 

\subsection{Comparison Methods}
We compare PinPoint against a particle filter (PF) baseline. To ensure a fair comparison, both methods use identical likelihood formulations, measurement inputs, and calibrated noise models.

The PF baseline propagates a set of particles to approximate the posterior over needle poses using stochastic sampling and resampling. Unlike SVN, which deterministically transports particles using gradient information, the PF relies on random propagation and resampling to maintain particle diversity. This baseline is conceptually similar to prior stereo needle tracking approaches based on particle filtering \cite{chiu_markerless_2022, ozguner_three-dimensional_2018}. 

Hyperparameters for each method were selected based on preliminary validation and held fixed across all experiments. For PinPoint, we used 50 SVN particles with up to 30 transport iterations initially to converge, reduced to 15 after the first 10 frames. Transport is terminated early when particle updates become sufficiently small and the maximum log-posterior value stalls across successive iterations. Specifically, transport is stopped once the 95th percentile of particle update magnitudes falls below fixed translational ($0.1$~mm) and rotational ($0.5^\circ$) thresholds, and the best log-posterior value changes by less than 0.02 for two consecutive iterations.

For the PF baseline, we use 3000 particles with a process model defined by the measured robot twist augmented with an isotropic random-walk component with standard deviations of $1$~mm and $5^\circ$ in translation and rotation, respectively. Systematic resampling is applied when the effective sample size falls below 30\% of the total number of particles.

\subsection{Calibration}
\label{sec: noise calibration}

Observation noise parameters were estimated empirically by evaluating each likelihood at the ground-truth needle pose across a calibration dataset. 
For each observation model, the sample mean and covariance were computed and used to whiten both the likelihoods and their Jacobians, producing approximately unit-variance error terms in the optimization.

\subsection{Ablation Design}
\label{sec: ablation}

Two types of internal ablation are performed to isolate the contribution of individual PinPoint components. 

First, we isolate the individual likelihoods to assess their contribution to accuracy and uncertainty. Using the full model with all likelihoods as a baseline, we first assessed the role of image-derived measurements by removing each component individually: the dense image likelihood $\mathcal{L}_dense$, the sparse image keypoint likelihood $\mathcal{L}_s$, and both image likelihood together, such that only robot-derived information remains.

Robot-based terms are similarly removed individually (the position and orientation components) and as a unit via $\mathcal{L}_{\mathrm{grasp}}$; when both are removed, the motion model is also disabled, leaving the estimator to rely  solely on monocular image geometry. This case is representative of free-needle  movement where the needle is not grasped by the robot at all.

Together, these ablations characterize how individual sensing and modeling components affect estimation accuracy, uncertainty quality, and posterior consistency in the proposed monocular framework. 

To isolate the contribution of Gauss-Newton preconditioning specifically, we compare PinPoint against a gradient-only SVGD baseline \cite{liu_stein_2016}. As noted in Sec. \ref{sec:svn_inference}, standard SVGD is recovered as a special case of the proposed SVN update when the per-particle Hessians are replaced by the identity and the kernel metric is set to isotropic with median-heuristic bandwidth. This baseline therefore shares the same likelihoods, particle count, and transport iterations as PinPoint, differing only in the use of first- versus second-order information.

\subsection{Evaluation Metrics}
Estimation accuracy is evaluated using translational and rotational pose errors measured with respect to ground truth needle poses. Let $T = (R,t)$ denote the estimated pose and $T_{\mathrm{gt}} = (R_{\mathrm{gt}}, t_{\mathrm{gt}})$ the corresponding ground truth pose.

Translational error is defined as the Euclidean distance between estimated and true needle origins:
\begin{equation}
e_t = \| t - t_{\mathrm{gt}} \|_2
\end{equation}

Rotational error is computed as the geodesic distance between orientations on $SO(3)$:
\begin{equation}
e_r = \cos^{-1}\!\left( \frac{\mathrm{trace}(R_{\mathrm{gt}}^T R) - 1}{2} \right)
\end{equation}

Because the proposed method explicitly represents a posterior distribution over needle poses, we additionally evaluate the quality and calibration of its uncertainty estimates. Uncertainty is quantified by the dispersion of particles in the tangent space, while consistency is assessed by comparing predicted uncertainty to observed estimation error over time.

Uncertainty metrics are computed separately for translation and rotation. For each component, we evaluate a negative log-likelihood score (NLL-S) as calibration metric that is distinct from the optimization objective $\mathcal{L}(T)$ of Section \ref{sec: problem}. This metric measures how probable the observed estimation error is under a Gaussian approximation of the predicted uncertainty, i.e., $\mathrm{err} \sim \mathcal{N}(0,\sigma^2)$ \cite{wursthorn2024uncertainty}. This Gaussian approximation is used solely for evaluation and does not imply a parametric posterior assumption. The NLL-S penalizes both large estimation errors and overconfident uncertainty predictions:

\begin{equation}
    \mathrm{NLL-S} = \tfrac{1}{2}\log(2\pi \sigma^2) + \tfrac{\mathrm{err}^2}{2\sigma^2}
\end{equation}

We additionally report the interval score, a scoring rule that jointly penalizes interval width and miscoverage, widely used for evaluating predictive uncertainty and calibration \cite{gneiting2007strictly}. Lower values indicate more accurate and better-calibrated uncertainty estimates:

\begin{gather}
    \mathrm{IS} =
(\mathrm{upp} - \mathrm{low})
+ \tfrac{2}{\alpha}\max(0, \mathrm{low} - \mathrm{err})\\
+ \tfrac{2}{\alpha}\max(0, \mathrm{err} - \mathrm{upp}) \notag
\end{gather}

\subsection{Multimodal Evaluation Criterion}
\label{sec:multimodal_eval}
To evaluate whether a solver (i.e., PinPoint and the PF) preserves multiple plausible pose hypotheses under ambiguous observations, we fit a Gaussian Mixture Model (GMM) to the particle distribution expressed in the tangent space of $SE(3)$. Using the MAP particle as a reference pose $ T_{\mathrm{ref}} = \hat{T}$, each particle $T^{(i)}$ is mapped to the tangent space via:
\begin{equation}
\xi^{(i)} = \log\!\left(T_{\mathrm{ref}}^{-1}T^{(i)}\right).
\end{equation}
and resulting vectors $\{\xi^{(i)}\}_{i=1}^N$ are normalized to a commensurable unit (1\,mm translation, 5\,$\degree$) before fitting. A one- and two-component GMM are fit via expectation-maximization (EM) and the Bayesian Information Criterion (BIC) selects between them. 

BIC selection alone, however, is not sufficient to conclude that a distribution is genuinely bimodal. EM is susceptible to degenerate solutions in which one component collapses onto a small subset of particles or captures noise rather than a true secondary mode \cite{mclachlan_finite_2000}. Such solutions can still achieve a lower BIC than the one-component fit if the likelihood improvement from fitting a narrow spike outweighs the parameter penalty. We therefore require that a two-component solution be physically meaningful according to Ashman's $D$ statistic \cite{ashman_detecting_1994}, which measures the separation between component means relative to their combined spread:
\begin{equation}
    D = \sqrt{2}\,\frac{|\mu_1 - \mu_2|}{\sqrt{\sigma_1^2 + \sigma_2^2}}
\end{equation}
where $\mu_k$ and $\sigma_k$ are the mean and standard deviation of the component $k$ along the axis connecting the two means. Ashman et al. showed that $D>2$ is necessary for the modes to be cleanly resolved as distinct peaks in the GMM. Below this threshold, the two Gaussians overlap such that the distribution remains effectively unimodal in appearance regardless of the BIC indicating a two-component fit. 

We additionally require that the minor component carry at least 10\% of the total mixture weight. Below this threshold the mixture is dominated at a 9:1 ratio by a single component, meaning the minor mode contributes negligibly to the posterior which shifts the mean by less than 10\% of the inter-mode separation and carrying insufficient weight to influence any downstream pose estimate or decision. This threshold is deliberately permissive: the rotational ambiguity that PinPoint is designed to represent produces asymmetric posteriors across most of the ambiguous configuration space, with one hypothesis naturally dominating as the needle rotates away from the maximally symmetric configuration. A conservative threshold captures genuine secondary modes across this full range while excluding only the degenerate near-zero weight solutions that reflect EM noise rather than distinct pose hypotheses.

A two-component fit is therefore accepted as \textit{genuinely bimodal} only when the BIC favors two components, $D>2$, and the minor component weight exceeds 10\% of the total. For analysis purposes, frames that satisfy all three of these criteria are labeled as Bimodal, all other frames are labeled as Unimodal regardless of BIC outcome. 

When a bimodal posterior is detected, we project the ground-truth pose $T_{gt}$ into the same normalized tangent-space coordinates and assign it to the nearest component mean. This enables us to assess not only whether a solver retains multiple hypotheses, but whether the `true mode' (i.e., the hypothesis with the greater weight) is consistent with the ground truth.

\begin{figure*}
    \centering
    \includegraphics[width=\textwidth]{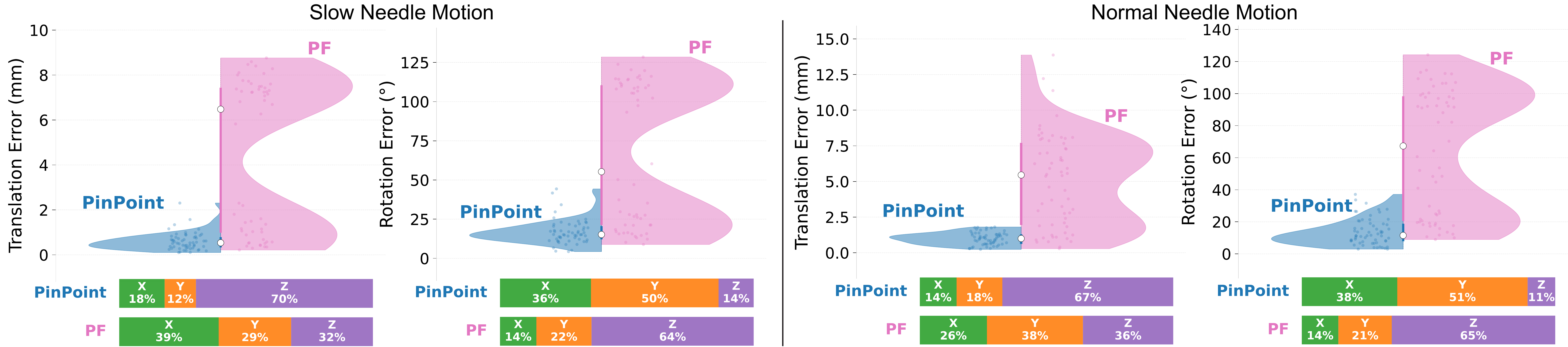}
    \caption{Distribution of translational and rotational pose estimation errors for PF and PinPoint across motion speeds. Each panel shows a half-violin with overlaid scatter for PF (pink, right) and PinPoint (blue, left), with white circles indicating the median. Stacked bars below each panel show the fractional contribution of each component axis (X: green, Y: orange, Z: purple) to total error. In both (Left) slow-speed trials and (Right) normal-speed trials, we show the rotational and translational errors.}
    \label{fig: svn pf results}
\end{figure*}

\section{Results and Discussion}

\subsection{Noise Calibration}
\begin{table}
\centering
\caption{Empirical standard deviations of observation likelihoods computed from calibration data.}

\begin{tabular}{l c}
\hline
Likelihood Type & Std. Dev. \\
\hline
Anchor (x, y, z) & $(0.41,\, 0.44,\, 1.00)$ mm \\
Perpendicular & $0.138$ \\
Conic & $1.235$ px \\
Keypoints (tip $u,v$; tail $u,v$) & $(1.006,\; 1.012;\; 0.921,\; 0.680)$ px \\
\hline
\end{tabular}
\label{tab:residual_calibration_std}
\end{table}

Table \ref{tab:residual_calibration_std} summarizes the resulting standard deviations. 
The robot position measurement exhibits anisotropic uncertainty, with sub-millimeter variance in x and y, and slightly larger variance along the depth axis. The robot orientation likelihood is tightly distributed.

Image-based measurements exhibit approximately pixel-level noise after calibration. 
The image keypoints remain close to unit-variance after whitening, indicating that the assumed observation model provides a reasonable approximation of the measurement statistics. 
The dense likelihood shows larger variance due to its sensitivity due to the segmentation boundary uncertainty.

Overall, the calibrated noise model produces balanced likelihood contributions across observation types, which improves numerical conditioning of the Gauss-Newton updates and reduces sensitivity to likelihood scaling.

\subsection{Tracking Comparison}

\begin{table*}
\centering
\caption{Comparison of PF and SVN tracking performance under slow and normal needle motion. Component-wise mean $\pm$ standard deviation and total error are reported for translation and rotation, along with 95th-percentile uncertainty (Unc), negative log-likelihood score (NLL-S), and interval score (IS). Lower is better for all metrics.}
\label{tab:pf_svn_comparison}
\vspace{0.4em}
\resizebox{\textwidth}{!}{%
\begin{tblr}{
  colspec = {l l l c c c c c c c},
  row{1} = {font=\bfseries, c},
  cell{2-Z}{1} = {font=\bfseries},
  vline{3,7,8} = {1-Z}{},
  hline{1,2,Z} = {-}{0.08em},
  hline{4,8} = {2-Z}{0.03em},
  hline{6} = {1-Z}{0.03em},
}
Method & Speed & Metric & X & Y & Z & Total & Unc (p95) & NLL-S & IS \\
PF  & Slow   & Trans (mm)
    & $2.81\pm2.63$ & $2.05\pm1.86$ & $2.31\pm1.76$ & $4.45\pm3.33$   & $8.42$   & $6.76$  & $6.62$ \\
    &        & Rot (deg)
    & $12.88\pm9.77$ & $19.51\pm12.00$ & $57.35\pm52.44$ & $65.99\pm44.74$ & $119.91$ & $79.10$ & $33.98$ \\
    & Normal   & Trans (mm)
    & $2.02\pm1.88$ & $2.89\pm1.92$ & $2.71\pm2.99$ & $5.04\pm3.26$   & $9.70$   & $6.87$  & $8.40$ \\
    &        & Rot (deg)
    & $12.93\pm9.75$ & $18.87\pm14.15$ & $57.78\pm42.27$ & $63.27\pm38.54$ & $112.56$  & $22.00$ & $30.94$ \\
PinPoint & Slow   & Trans (mm)
    & $0.14\pm0.13$ & $0.10\pm0.09$ & $0.55\pm0.40$ & $\mathbf{0.60\pm0.39}$ & $\mathbf{1.20}$ & $\mathbf{1.86}$ & $\textbf{2.426}$ \\
    &        & Rot (deg)
    & $8.62\pm6.92$ & $12.09\pm8.19$ & $3.32\pm1.97$ & $16.92\pm7.74$ & $30.80$ & $\textbf{4.82}$ & $\textbf{24.15}$ \\
    & Normal   & Trans (mm)
    & $0.20\pm0.13$ & $0.25\pm0.18$ & $0.92\pm0.44$ & $1.00\pm0.43$ & $1.75$ & $1.79$ & $11.18$ \\
    &        & Rot (deg)
    & $7.46\pm5.30$ & $10.18\pm8.24$ & $2.15\pm1.40$ & $\mathbf{13.80\pm8.23}$ & $\mathbf{28.02}$ & $4.98$ & $32.14$ \\
\end{tblr}%
}
\end{table*}

Quantitative results comparing PinPoint to the particle filter (PF) baseline are summarized in Table \ref{tab:pf_svn_comparison} and Fig. \ref{fig: svn pf results}, reported separately for each motion condition.

Under slow motion compared to the PF, PinPoint reduces mean translational error by 87\% (PF=4.45\,mm to PinPoint=0.60\,mm) and mean rotational error by 74\% (PF=65.99\,$\degree$\ to PinPoint=16.92\,$\degree$). The advantage is maintained under normal-speed motion, where PinPoint achieves 1.00\,mm and 13.80\,$\degree$\ compared to the PF's 5.04\,mm and 63.27\,$\degree$, an 80\% reduction in translational error and a 78\% reduction in rotational error. Notably, PinPoint's rotational error remains nearly constant across motion speeds (16.92\,$\degree$\ slow vs.\ 13.80\,$\degree$\ normal), while the PF's translational error grows and its rotational error stays large and variable. This robustness is consistent with SVN's deterministic transport avoiding the weight collapse that degrades importance-weighted filters under fast motion, and with the robot-grasp constraints providing continuous 3D grounding independent of visual tracking quality.

A characteristic pattern emerges in the component-wise breakdown of PinPoint's errors that reflects the geometry of the monocular estimation problem. Translation error is dominated by the $z$ component (0.55\,mm slow, 0.92\,mm normal), which corresponds to depth along the camera optical axis—the direction most ambiguous under monocular projection. Similarly, rotation error is largest about $x$ and $y$ (tip-tilt axes), which corresponds to out-of-plane orientation that monocular appearance provides only weak cues for, while the $z$ rotation component (needle roll about its own axis) is substantially smaller (3.32\,$\degree$\ slow, 2.15\,$\degree$\ normal). This demonstrates how PinPoint concentrates likelihood uncertainty precisely where the image provides the least information, while remaining highly accurate along the well-constrained axes.

Uncertainty calibration follows a similar pattern. Under slow motion, PinPoint achieves a 3.6$\times$ reduction in translational NLL-S (1.86 vs.\ 6.76) and a 16.4$\times$ reduction in rotational NLL-S (4.82 vs.\ 79.10) relative to PF, with tighter and better-calibrated uncertainty bounds across both axes. Under normal-speed motion, PinPoint again achieves substantially better rotational uncertainty calibration, with a 4.4$\times$ reduction in rotational NLL-S (4.98 vs.\ 22.00), reflecting the PF's difficulty maintaining a calibrated rotational posterior as speed increases. Translational NLL-S at normal speed favors PinPoint as well (1.79 vs.\ 6.87), in contrast to the prior dataset where PF's wider intervals artificially reduced its NLL-S penalty. The IS results are mixed: PinPoint's sharper posteriors are penalized on IS when the target occasionally falls outside tight bounds, particularly under slow motion where the PF's wider intervals incidentally achieve better coverage.



\subsection{Ablation Study}
\label{sec:ablation}

\begin{figure}
    \centering
    \includegraphics[width=\columnwidth]{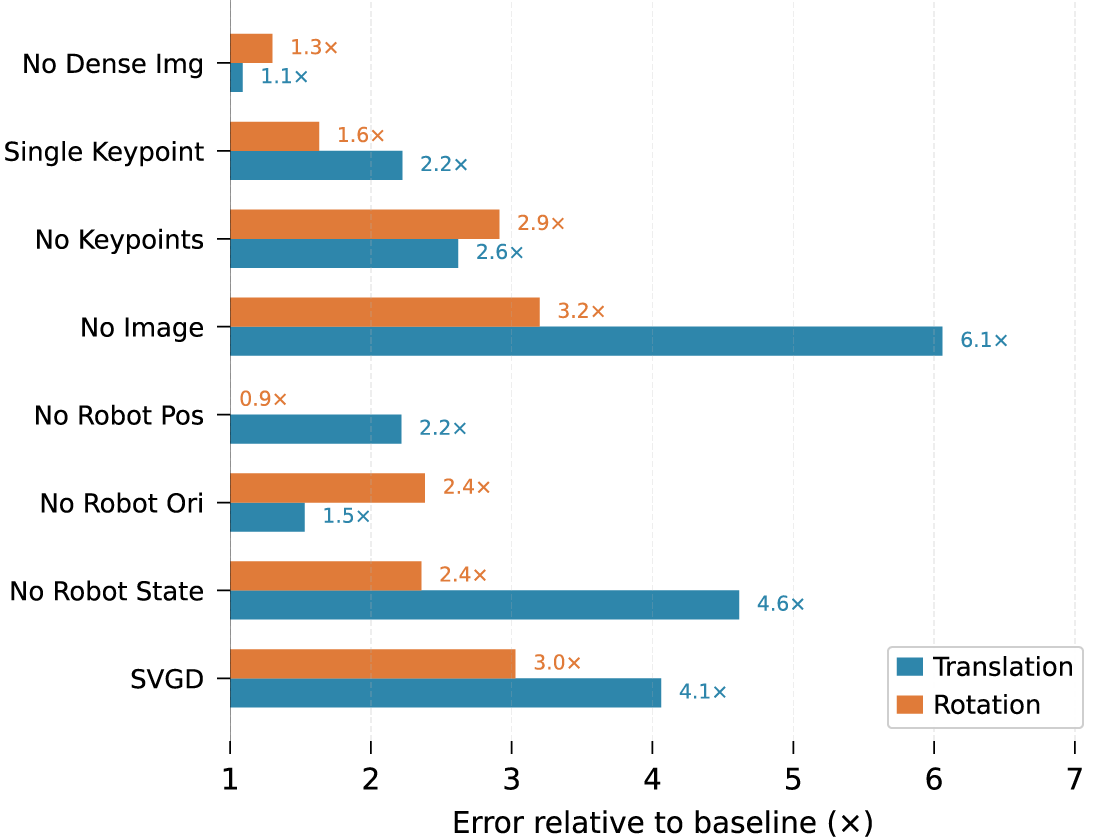}
    \caption{A bar graph breakdown of the error degradation about the translational and rotational under the given ablation.}
    \label{fig: ablation_bars}
\end{figure}

\begin{table*}
\centering
\caption{Ablation study evaluating the contribution of individual likelihood terms to needle pose estimation accuracy using the SVN method. Mean $\pm$ std error is reported along with 95\% uncertainty coverage (Unc), negative log-likelihood score (NLL-S), and interval score (IS). Lower is better for all metrics.}
\vspace{0.5em}

\begin{tblr}{
  colspec = {l c c c c c c c c},
  cell{1}{1} = {r=2}{},
  cell{1}{2} = {c=4}{c},
  cell{1}{6} = {c=4}{c},
  vline{2,6} = {1-Z}{},
  hline{1,3,Z} = {-}{0.08em},
}
Ablation & Translation (mm) & & & & Rotation (deg) & & & \\
 & Mean $\pm$ Std & Unc & NLL-S & IS & Mean $\pm$ Std & Unc & NLL-S & IS \\

\textbf{All (Full Model)} &
0.75$\pm$0.37 & 1.37 & 1.40 & 3.03 &
9.87$\pm$4.60 & 18.29 & 4.82 & 24.15 \\

No Dense Image &
0.81$\pm$0.48 & 1.80 & 1.49 & 4.36 &
12.83$\pm$8.95 & 31.55 & 5.43 & 40.06 \\

1 Sparse Keypoint &
1.66$\pm$1.80 & 3.94 & 3.13 & 4.98 &
16.10$\pm$13.40 & 47.49 & 22.13 & 45.99 \\

No Sparse Keypoints &
1.96$\pm$1.72 & 4.30 & 4.57 & 5.87 &
28.73$\pm$21.80 & 60.52 & 52.02 & 58.49 \\

No Image &
4.53$\pm$2.41 & 10.14 & 9.51 & 5.98 &
31.55$\pm$43.06 & 143.99 & 11.49 & 59.53 \\

No Robot Position &
1.66$\pm$2.03 & 4.37 & 2.46 & 6.35 &
9.26$\pm$8.51 & 21.37 & 4.15 & 29.62 \\

No Robot Orientation &
1.14$\pm$0.78 & 2.55 & 1.99 & 5.46 &
23.52$\pm$10.18 & 41.96 & 4.96 & 104.34 \\

No Robot \& No Motion Model &
3.45$\pm$3.86 & 9.18 & 6.37 & 12.58 &
23.26$\pm$23.30 & 52.29 & 7.20 & 108.33 \\

SVGD (No Hessian) &
1.52$\pm$0.64 & 6.00 & 6.89 & 6.76 &
29.20$\pm$6.40 & 48.10 & 4.90 & 118.87 \\

\end{tblr}
\label{tab: ablation}
\end{table*}

The results of the ablation study, shown in Table \ref{tab: ablation}, highlight both complementary sensing roles and partial redundancy across likelihoods, and using the slow-motion dataset from the previous section. Removing image-based measurements produces the largest degradation in performance. In particular, eliminating sparse keypoints substantially increases both translational and rotational error, while removing all image information results in the highest overall error and uncertainty. This behavior is expected, as monocular image observations provide the primary geometric constraints on the needle shape and orientation. Figure \ref{fig: ablation_bars} illustrates a bar graph with the relative degredation of each ablation. 

Dense segmentation contributes more modestly when sparse keypoints are available, but improves robustness and uncertainty calibration, particularly in rotational estimates. This suggests that the dense likelihood acts primarily as a stabilizing global shape constraint rather than a dominant pose constraint.

Grasp likelihoods play a complementary role by anchoring the solution in 3D space. Removing the robot position likelihood increases translational error, particularly along the z-direction, while removing the robot orientation constraint significantly increases rotational error, reflecting the loss of the grasp-imposed rotational constraint. When both robot constraints and the motion model are removed, the estimator relies solely on monocular image geometry, resulting in degraded accuracy and substantially inflated uncertainty.

\begin{figure*}
    \centering
    \includegraphics[width=\textwidth]{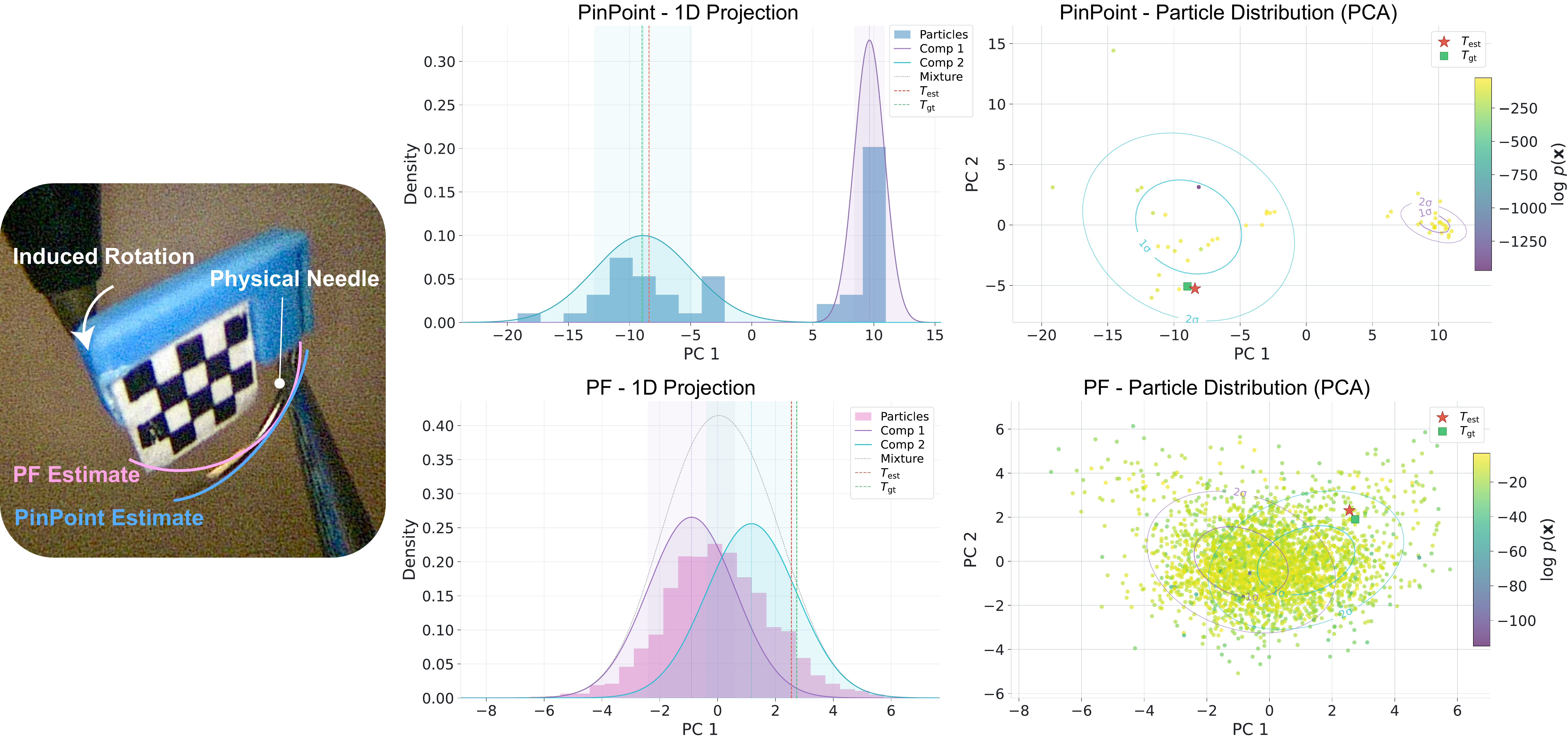}
    \caption{PinPoint correctly resolves pose ambiguity that confounds the particle filter. (Left) A suturing needle undergoing rotation presents a visually symmetric appearance, inducing a bimodal posterior over needle pose. PinPoint and PF estimates are shown overlaid on the endoscopic image. (Right, Top Row) PinPoint represents the posterior as a two-component GMM with well-separated modes in PCA space; the estimate $T_{est}$ is located within the same mode as the ground truth $T_{gt}$, while correctly maintaining the alternative hypothesis. (Right, Bottom Row) The PF fits a two-component mixture but fails to separate the hypotheses, as both components heavily overlap in PCA space, producing a diffuse, ambiguous distribution.}
    \label{fig:multimodal_clustering}
\end{figure*}

Compared to SVGD, PinPoint reduces mean translational error by 75.4\%  and mean rotational error by 66.9\%. The condition number $\kappa(H)$ ranges from $8.99 \times 10^{6}$ to $1.65 \times 10^{11}$ with a median of $1.05 \times 10^{7}$, reflecting the anisotropic observability of monocular pose: depth and out-of-plane rotation are poorly constrained while in-plane translation and rotation are well-constrained.

Overall, the ablation results confirm that accurate monocular needle tracking requires combining image-based geometric information with robot-grasp constraints. No single likelihood is sufficient on its own, but together they provide complementary constraints that improve both accuracy and uncertainty.

\subsection{Multimodal Evaluation}

We apply the bimodality criterion of Sec. \ref{sec:multimodal_eval} independently to the SVN and PF distributions. PinPoint's posterior was classified as Bimodal in 83.6\% of frames, compared to 28.9\% for the PF baseline. The substantially lower bimodality rate of the PF baseline is consistent with the known tendency of importance-weighted filters to undergo weight collapse under sharply peaked likelihoods \cite{johansen2009tutorial}, as particles cluster near a single high-likelihood region; the effective sample size drops and the distribution loses its capacity to maintain separated modes.

In 62\% of all frames, PinPoint correctly maintained a bimodal posterior while the PF had already collapsed. On these frames, the PF pvodies a confident point estimate with no representation of the alternative hypothesis. The most consequential failure mode for downstream planning is where an undetected alternative pose hypothesis could lead to an incorrect action.

Beyond detecting bimodality, PinPoint correctly weights the two hypotheses: in 100\% of bimodal frames where ground-truth pose could be assigned to a component, the ground-truth fell within the dominant mode (the hypothesis carrying greater mixture weight). The PF achieved this alignment in only 60\% of frames, indicating that even when the PF maintains two components, it frequently assigns greater weight to the incorrect hypothesis.

The distributional difference is also reflected in uncertainty calibration. Evaluated under the fitted posterior density, PinPoint achieves a mean rotational NLL-S of 4.94 compared to 8.66 for the PF baseline, a 43.0\% reduction, indicating that PinPoint's probability mass is better distributed relative to the ground-truth pose. The improvement in translational NLL-S is modest (2.60 vs.\ 2.73), consistent with translation being less affected by the out-of-plane rotation ambiguity.

Figure \ref{fig:multimodal_clustering} shows a representative ambiguous frame. The PinPoint particle cloud (top) exhibits two well-separated clusters in the principal tangent-space directions  (D=4.9), with fitted GMM ellipses and $T_{gt}$. The PF cloud (bottom) has collapsed to a single cluster inconsistent with $T_{gt}$, receiving a Unimodal verdict.

\subsection{Suturing Demonstration}

\begin{figure*}
    \centering
    \includegraphics[width=\textwidth]{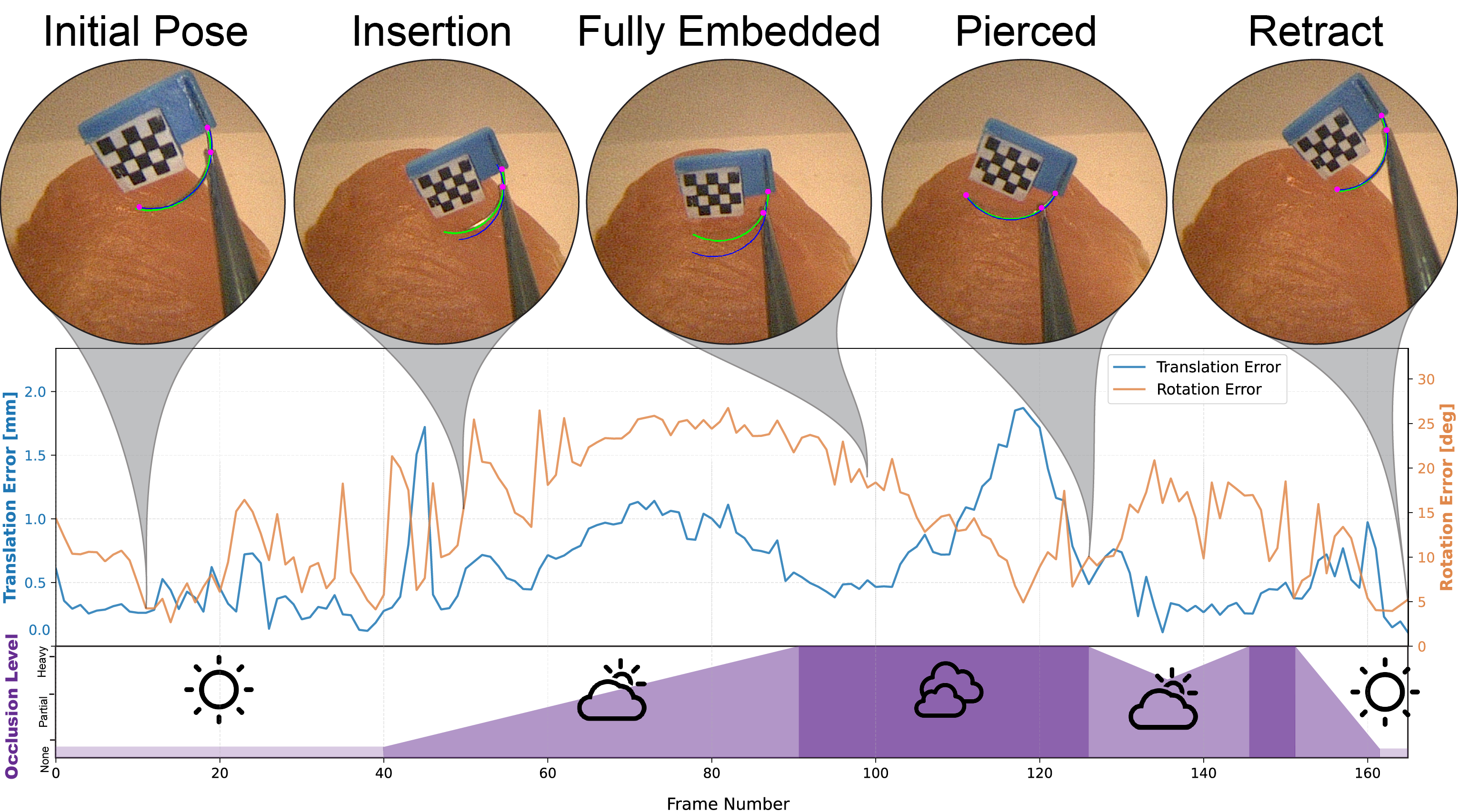}
    \caption{Graph of translation (blue) and rotation (orange) errors during suturing, with representative images shown at key frames. A graph of the occlusion level is graphed below, with icons illustrating the levels of none, partial, and heavy occlusion. The graph illustrates how occlusion increases during piercing until the tip emerges on the other side of the tissue. Occlusion then decreases as the needle is retracted until it fully exits the tissue.}
    \label{fig:suturing error plot}
\end{figure*}

A representative example of translational and rotational error over time is shown in Fig.~\ref{fig:suturing error plot}. 
Across suturing trials, mean translational error before the pierce was 0.52\,mm and mean rotational error 13.82\degree, demonstrating stable tracking performance during tissue interaction. For each trial, about 50\% of the time the needle was at least partially occluded during piercing, in which the image observations were intermittently unavailable. Under occlusion, on average, the translation error was 1.34\,mm and rotation error to 19.18\degree. 

As the needle enters the tissue, the visible backbone segmentation gradually disappears and the needle tip keypoint is unobservable, corresponding to a partial occlusion. During this interval, estimation relies primarily on robot-based grasp constraints, the motion model, and prior particle distribution. This partial occlusion leads to increased uncertainty, as shown in Fig. \ref{fig:suturing error plot}. When the needle is fully embedded, only the tail keypoint remains visible, corresponding to a heavy occlusion, leading to the largest estimation errors observed during the task. This error increase is most pronounced in the translational component. The rigid couping assumption between the robot motion and needle motion begins to breakdown; when the needle is embedded in the tissue, the robot may deflect or slip, resulting in the needle remaining stationary while the robot is believed to have moved.

Importantly, the estimator remains stable throughout these occlusion events. When the needle tip re-emerges, additional image constraints become available and the particle distribution rapidly contracts toward the correct pose. Once the needle is fully visible again, the estimate returns to its nominal accuracy. 

These results highlight the complementary roles of image-based and robot-based measurements in PinPoint. Image observations provide strong geometric constraints when visible, while robot kinematics and grasp constraints maintain estimator stability during periods of reduced visual information. This behavior is particularly important for robotic suturing, where temporary occlusions during needle insertion are unavoidable.

\section{Conclusion}
We presented PinPoint, a probabilistic variational inference framework for real-time monocular needle pose estimation during robotic suturing. PinPoint formulates needle tracking as a multimodal inference problem over SE(3), combining geometric image likelihoods and robot-based grasp constraints within a Stein Variational Newton framework that converges to correct posterior modes while preserving hypothesis diversity through kernel-based repulsion.


Experiments across controlled motion, induced out-of-plane rotations, and suturing in ex vivo tissue validate the approach. PinPoint achieves 1.00\,mm mean translational error and 13.80\,$\degree$ mean rotational error under normal-speed motion and remains stable under heavy occlusions with mean errors of 1.34\,mm and 19.18\,$\degree$. Ablation studies confirm the complementary roles of image- and robot-derived likelihoods. On induced-rotation sequences, PinPoint maintains a bimodal posterior in 93\% of frames compared to 41\% for the particle filter baseline, demonstrating that the framework correctly represents geometric ambiguity rather than suppressing it.


Together, these results establish that probabilistic geometric inference with robot-grasp constraints enable reliable monocular needle tracking across the full range of conditions encountered in robotic suturing. 
Future work will integrate PinPoint with uncertainty-aware autonomous suturing planners, using the posterior distribution to gate decisions such as insertion timing, re-grasp requests, and tissue engagement.
The second-order Hessian structure offers an active sensing capability: the least-constrained eigenvector identifies directions for needle repositioning to reduce uncertainty before critical actions. Together, these extensions point toward a suturing autonomy pipeline that is aware of and can actively reduce risk at each step while advancing the suturing task.

\bibliographystyle{ieeetr}
\bibliography{ManualRefs, ZoteroRefs}

@inproceedings{dalmeida_stent_2026,
  author    = {d'Almeida, Jesse F. and Branscombe, Lauren and Chara, Alejandro O. and Stern, Susheela Sharma and Webster III, Robert J.},
  title     = {Needle-and-Thread Suturing With Concentric Tube Robots for Tracheal Stent Fixation},
  booktitle = {2026 {International} {Symposium} on {Medical} {Robotics} ({ISMR})},
  year      = {2026},
  address   = {Nashville, TN, USA},
  publisher = {IEEE},
  note = {Accepted for publication},
}

@article{dalmeida_prostatectomy_2026,
  author  = {d'Almeida, Jesse F. and Maciolek, Kimberly A. and Wilke, Ethan and Shrand, Jason A. and Shepard, Lauren and Ertop, Tayfun E. and Ghazi, Ahmed E. and Webster III, Robert J. and Herrell III, S. Duke and Kavoussi, Nicholas L.},
  title   = {Robotic Transurethral Vesicourethral Anastomosis after Radical Prostatectomy in Validated Models},
  journal = {Journal of Endourology},
  year    = {2025},
  note = {Accepted for publication},
}

@book{mclachlan_finite_2000,
  author    = {McLachlan, Geoffrey J. and Peel, David},
  title     = {Finite Mixture Models},
  publisher = {Wiley},
  year      = {2000}
}

@article{ashman_detecting_1994,
  author  = {Ashman, Keith M. and Bird, Caleb M. and Zepf, Stephen E.},
  title   = {Detecting Bimodality in Astronomical Datasets},
  journal = {The Astronomical Journal},
  volume  = {108},
  pages   = {2348--2361},
  year    = {1994}
}

@article{johansen2009tutorial,
  title={A tutorial on particle filtering and smoothing: Fifteen years later},
  author={Johansen, Adam},
  year={2009}
}

@book{faraz2000engineering,
  title={Engineering approaches to mechanical and robotic design for minimally invasive surgery (MIS)},
  author={Faraz, Ali and Payandeh, Shahram},
  volume={545},
  year={2000},
  publisher={Springer Science \& Business Media}
}

@manual{virtuoso-ves-manual,
  title        = {{Virtuoso} Endoscopy System},
  organization = {Virtuoso Surgical},
  address      = {Nashville, TN},
  note         = {\url{https://virtuososurgical.net/technology/}. Accessed 2025-09-15}
}

@inproceedings{isik2026benchmark,
  author    = {Isik, Dilara and Oguine, Kanyifeechukwu and DiSanto, Nick and d'Almeida, Jesse and Webster, Robert J., III and Oguz, Ipek and Li, Hao},
  title     = {A benchmark study of methods for surgical instrument segmentation in central airway endoscopy},
  booktitle = {Medical Imaging 2026: Clinical and Biomedical Imaging},
  volume    = {13925},
  pages     = {13925-89},
  year      = {2026},
  month     = {February},
  publisher = {SPIE},
  address   = {San Francisco, California, USA}
}

@inproceedings{wehrbein2021probabilistic,
  title={Probabilistic monocular 3d human pose estimation with normalizing flows},
  author={Wehrbein, Tom and Rudolph, Marco and Rosenhahn, Bodo and Wandt, Bastian},
  booktitle={Proceedings of the IEEE/CVF international conference on computer vision},
  pages={11199--11208},
  year={2021}
}

@article{opencv_library,
    author = {Bradski, G.},
    journal = {Dr. Dobb's Journal of Software Tools},
    title = {{The OpenCV Library}},
    year = {2000}
}

@inproceedings{barragan2024realistic,
  title={Realistic data generation for 6d pose estimation of surgical instruments},
  author={Barragan, Juan Antonio and Zhang, Jintan and Zhou, Haoying and Munawar, Adnan and Kazanzides, Peter},
  booktitle={2024 IEEE international conference on robotics and automation (ICRA)},
  pages={13347--13353},
  year={2024},
  organization={IEEE}
}

@inproceedings{zheng2008another,
  title={Another way of looking at monocular circle pose estimation},
  author={Zheng, Yinqiang and Ma, Wenjuan and Liu, Yuncai},
  booktitle={2008 15th IEEE International Conference on Image Processing},
  pages={861--864},
  year={2008},
  organization={IEEE}
}

@book{nocedal2006numerical,
  title={Numerical optimization},
  author={Nocedal, Jorge and Wright, Stephen J},
  year={2006},
  publisher={Springer}
}

@book{hartley2003multiple,
  title={Multiple view geometry in computer vision},
  author={Hartley, Richard and Zisserman, Andrew},
  year={2003},
  publisher={Cambridge university press}
}

@inproceedings{bui20206,
  title={6d camera relocalization in ambiguous scenes via continuous multimodal inference},
  author={Bui, Mai and Birdal, Tolga and Deng, Haowen and Albarqouni, Shadi and Guibas, Leonidas and Ilic, Slobodan and Navab, Nassir},
  booktitle={European Conference on Computer Vision},
  pages={139--157},
  year={2020},
  organization={Springer}
}

@inproceedings{manhardt2019explaining,
  title={Explaining the ambiguity of object detection and 6d pose from visual data},
  author={Manhardt, Fabian and Arroyo, Diego Martin and Rupprecht, Christian and Busam, Benjamin and Birdal, Tolga and Navab, Nassir and Tombari, Federico},
  booktitle={Proceedings of the IEEE/CVF International Conference on Computer Vision},
  pages={6841--6850},
  year={2019}
}

@article{gneiting2007strictly,
  title={Strictly proper scoring rules, prediction, and estimation},
  author={Gneiting, Tilmann and Raftery, Adrian E},
  journal={Journal of the American statistical Association},
  volume={102},
  number={477},
  pages={359--378},
  year={2007},
  publisher={Taylor \& Francis}
}

@article{wursthorn2024uncertainty,
  title={Uncertainty quantification with deep ensembles for 6d object pose estimation},
  author={Wursthorn, Kira and Hillemann, Markus and Ulrich, Markus},
  journal={arXiv preprint arXiv:2403.07741},
  year={2024}
}

@misc{blanco-claraco_tutorial_2022,
	title = {A tutorial on {SE}(3) transformation parameterizations and on-manifold optimization},
	url = {http://arxiv.org/abs/2103.15980},
	doi = {10.48550/arXiv.2103.15980},
	urldate = {2025-10-30},
	publisher = {arXiv},
	author = {Blanco-Claraco, José Luis},
	month = apr,
	year = {2022},
}

@article{pedram_autonomous_2021,
	title = {Autonomous suturing framework and quantification using a cable-driven surgical robot},
	volume = {37},
	issn = {1941-0468},
	url = {https://ieeexplore.ieee.org/abstract/document/9247988?casa_token=V0RqWnNSLFQAAAAA:OSlM8i-Hsa5lDvyLw3a7YBpSNbi2IDHWrpj2Q45V1A7Gue5-_M5isB3CM7NTuqF1CC8VInB0IDM},
	doi = {10.1109/TRO.2020.3031236},
	number = {2},
	urldate = {2025-03-04},
	journal = {IEEE Transactions on Robotics},
	author = {Pedram, Sahba Aghajani and Shin, Changyeob and Ferguson, Peter Walker and Ma, Ji and Dutson, Erik P. and Rosen, Jacob},
	year = {2021},
	pages = {404--417},
}

@inproceedings{sen_automating_2016,
	title = {Automating multi-throw multilateral surgical suturing with a mechanical needle guide and sequential convex optimization},
	isbn = {978-1-4673-8026-3},
	url = {http://ieeexplore.ieee.org/document/7487611/},
	doi = {10.1109/ICRA.2016.7487611},
	language = {en},
	urldate = {2025-08-25},
	booktitle = {2016 {IEEE} {International} {Conference} on {Robotics} and {Automation} ({ICRA})},
	publisher = {IEEE},
	author = {Sen, Siddarth and Garg, Animesh and Gealy, David V. and McKinley, Stephen and Jen, Yiming and Goldberg, Ken},
	year = {2016},
	pages = {4178--4185},
}

@inproceedings{ozguner_visually_2021,
	title = {Visually guided needle driving and pull for autonomous suturing},
	url = {https://ieeexplore.ieee.org/abstract/document/9551453},
	doi = {10.1109/CASE49439.2021.9551453},
	urldate = {2025-10-05},
	booktitle = {2021 {IEEE} 17th {International} {Conference} on {Automation} {Science} and {Engineering} ({CASE})},
	author = {Özgüner, Orhan and Shkurti, Tom and Lu, Su and Newman, Wyatt and Çavuşoğlu, M. Cenk},
	year = {2021},
	note = {ISSN: 2161-8089},
	pages = {242--248},
}

@article{webster_iii_design_2010,
	title = {Design and kinematic modeling of constant curvature continuum robots: a review},
	volume = {29},
	issn = {0278-3649},
	shorttitle = {Design and kinematic modeling of constant curvature continuum robots},
	url = {https://doi.org/10.1177/0278364910368147},
	doi = {10.1177/0278364910368147},
	language = {en},
	number = {13},
	urldate = {2024-08-27},
	journal = {The International Journal of Robotics Research},
	author = {Webster III, Robert J. and Jones, Bryan A.},
	year = {2010},
	pages = {1661--1683},
}

@article{hubens_performance_2003,
	title = {A performance study comparing manual and robotically assisted laparoscopic surgery using the da {Vinci} system},
	volume = {17},
	issn = {1432-2218},
	url = {https://doi.org/10.1007/s00464-002-9248-1},
	doi = {10.1007/s00464-002-9248-1},
	language = {en},
	number = {10},
	urldate = {2025-09-30},
	journal = {Surgical Endoscopy And Other Interventional Techniques},
	author = {Hubens, G. and Coveliers, H. and Balliu, L. and Ruppert, M. and Vaneerdeweg, W.},
	year = {2003},
	pages = {1595--1599},
}

@article{li_monocular_2024,
	title = {On the {Monocular} 3-{D} {Pose} {Estimation} for {Arbitrary} {Shaped} {Needle} in {Dynamic} {Scenes}: {An} {Efficient} {Visual} {Learning} and {Geometry} {Modeling} {Approach}},
	volume = {6},
	copyright = {https://ieeexplore.ieee.org/Xplorehelp/downloads/license-information/IEEE.html},
	issn = {2576-3202},
	shorttitle = {On the {Monocular} 3-{D} {Pose} {Estimation} for {Arbitrary} {Shaped} {Needle} in {Dynamic} {Scenes}},
	url = {https://ieeexplore.ieee.org/document/10472639/},
	doi = {10.1109/TMRB.2024.3377357},
	language = {en},
	number = {2},
	urldate = {2026-01-21},
	journal = {IEEE Transactions on Medical Robotics and Bionics},
	author = {Li, Bin and Lu, Bo and Lin, Hongbin and Wang, Yaxiang and Zhong, Fangxun and Dou, Qi and Liu, Yun-Hui},
	month = may,
	year = {2024},
	pages = {460--474},
}

@inproceedings{detommaso_stein_2018,
	title = {A {Stein} variational {Newton} method},
	volume = {31},
	url = {https://proceedings.neurips.cc/paper_files/paper/2018/hash/fdaa09fc5ed18d3226b3a1a00f1bc48c-Abstract.html},
	urldate = {2025-12-24},
	booktitle = {Advances in {Neural} {Information} {Processing} {Systems}},
	publisher = {Curran Associates, Inc.},
	author = {Detommaso, Gianluca and Cui, Tiangang and Marzouk, Youssef and Spantini, Alessio and Scheichl, Robert},
	year = {2018},
}

@inproceedings{liu_stein_2016,
	title = {Stein {Variational} {Gradient} {Descent}: {A} {General} {Purpose} {Bayesian} {Inference} {Algorithm}},
	volume = {29},
	shorttitle = {Stein {Variational} {Gradient} {Descent}},
	url = {https://proceedings.neurips.cc/paper/2016/hash/b3ba8f1bee1238a2f37603d90b58898d-Abstract.html},
	urldate = {2025-12-24},
	booktitle = {Advances in {Neural} {Information} {Processing} {Systems}},
	publisher = {Curran Associates, Inc.},
	author = {Liu, Qiang and Wang, Dilin},
	year = {2016},
}

@article{eade_lie_nodate,
	title = {Lie {Groups} for {2D} and {3D} {Transformations}},
	language = {en},
	author = {Eade, Ethan},
}

@inproceedings{cross_quadric_1998,
	title = {Quadric reconstruction from dual-space geometry},
	url = {https://ieeexplore.ieee.org/abstract/document/710697},
	doi = {10.1109/ICCV.1998.710697},
	urldate = {2025-10-30},
	booktitle = {Sixth {International} {Conference} on {Computer} {Vision}},
	author = {Cross, G. and Zisserman, A.},
	month = jan,
	year = {1998},
	pages = {25--31},
}

@article{barfoot_associating_2014,
	title = {Associating {Uncertainty} {With} {Three}-{Dimensional} {Poses} for {Use} in {Estimation} {Problems}},
	volume = {30},
	issn = {1941-0468},
	url = {https://ieeexplore.ieee.org/document/6727494/},
	doi = {10.1109/TRO.2014.2298059},
	number = {3},
	urldate = {2025-10-16},
	journal = {IEEE Transactions on Robotics},
	author = {Barfoot, Timothy D. and Furgale, Paul T.},
	month = jun,
	year = {2014},
	pages = {679--693},
}

@inproceedings{iyer_single_2013,
	title = {A single arm, single camera system for automated suturing},
	url = {https://ieeexplore.ieee.org/abstract/document/6630582},
	doi = {10.1109/ICRA.2013.6630582},
	urldate = {2025-10-18},
	booktitle = {2013 {IEEE} {International} {Conference} on {Robotics} and {Automation}},
	author = {Iyer, Santosh and Looi, Thomas and Drake, James},
	month = may,
	year = {2013},
	note = {ISSN: 1050-4729},
	pages = {239--244},
}

@article{wang_monocular_2025,
	title = {Monocular suture needle pose detection using synthetic data augmented convolutional neural network},
	volume = {20},
	issn = {1861-6429},
	url = {https://doi.org/10.1007/s11548-025-03467-1},
	doi = {10.1007/s11548-025-03467-1},
	language = {en},
	number = {10},
	urldate = {2025-10-18},
	journal = {International Journal of Computer Assisted Radiology and Surgery},
	author = {Wang, Yifan and Perez, Saul Alexis Heredia and Harada, Kanako},
	month = oct,
	year = {2025},
	pages = {2019--2030},
}

@misc{chiu_real-time_2022,
	title = {Real-{Time} {Constrained} {6D} {Object}-{Pose} {Tracking} of {An} {In}-{Hand} {Suture} {Needle} for {Minimally} {Invasive} {Robotic} {Surgery}},
	url = {http://arxiv.org/abs/2210.11973},
	doi = {10.48550/arXiv.2210.11973},
	urldate = {2025-10-18},
	publisher = {arXiv},
	author = {Chiu, Zih-Yun and Richter, Florian and Yip, Michael C.},
	month = oct,
	year = {2022},
	note = {arXiv:2210.11973 [cs]},
}

@inproceedings{chiu_markerless_2022,
	title = {Markerless {Suture} {Needle} {6D} {Pose} {Tracking} with {Robust} {Uncertainty} {Estimation} for {Autonomous} {Minimally} {Invasive} {Robotic} {Surgery}},
	url = {https://ieeexplore.ieee.org/abstract/document/9981966},
	doi = {10.1109/IROS47612.2022.9981966},
	urldate = {2025-10-05},
	booktitle = {2022 {IEEE}/{RSJ} {International} {Conference} on {Intelligent} {Robots} and {Systems} ({IROS})},
	author = {Chiu, Zih-Yun and Liao, Albert Z and Richter, Florian and Johnson, Bjorn and Yip, Michael C.},
	month = oct,
	year = {2022},
	note = {ISSN: 2153-0866},
	pages = {5286--5292},
}

@inproceedings{ozguner_three-dimensional_2018,
	title = {Three-{Dimensional} {Surgical} {Needle} {Localization} and {Tracking} {Using} {Stereo} {Endoscopic} {Image} {Streams}},
	url = {https://ieeexplore.ieee.org/abstract/document/8460867},
	doi = {10.1109/ICRA.2018.8460867},
	urldate = {2025-10-03},
	booktitle = {2018 {IEEE} {International} {Conference} on {Robotics} and {Automation} ({ICRA})},
	author = {Özgüner, Orhan and Hao, Ran and Jackson, Russell C. and Shkurti, Tom and Newman, Wyatt and Cavusoglu, M. Cenk},
	month = may,
	year = {2018},
	note = {ISSN: 2577-087X},
	pages = {6617--6624},
}

@article{shkurti_systematic_2025,
	title = {A {Systematic} {Review} of {Task} {Automation} in {Surgical} {Robotics}},
	volume = {7},
	issn = {2576-3202},
	url = {https://ieeexplore.ieee.org/stampPDF/getPDF.jsp},
	doi = {10.1109/TMRB.2025.3583182},
	number = {3},
	urldate = {2025-08-28},
	journal = {IEEE Transactions on Medical Robotics and Bionics},
	author = {Shkurti, Thomas E. and Cenk Çavuşoğlu, M.},
	month = aug,
	year = {2025},
	pages = {863--880},
}

@inproceedings{wilcox_learning_2022,
	title = {Learning to {Localize}, {Grasp}, and {Hand} {Over} {Unmodified} {Surgical} {Needles}},
	url = {https://ieeexplore.ieee.org/document/9812393/?arnumber=9812393},
	doi = {10.1109/ICRA46639.2022.9812393},
	urldate = {2025-03-10},
	booktitle = {2022 {International} {Conference} on {Robotics} and {Automation} ({ICRA})},
	author = {Wilcox, Albert and Kerr, Justin and Thananjeyan, Brijen and Ichnowski, Jeffrey and Hwang, Minho and Paradis, Samuel and Fer, Danyal and Goldberg, Ken},
	month = may,
	year = {2022},
	pages = {9637--9643},
}

@article{harvey_novel_2020,
	title = {A {Novel} {Robotic} {Endoscopic} {Device} {Used} for {Operative} {Hysteroscopy}},
	volume = {27},
	issn = {1553-4650, 1553-4669},
	url = {https://www.webofscience.com/api/gateway?GWVersion=2&SrcAuth=DynamicDOIArticle&SrcApp=WOS&KeyAID=10.1016%2Fj.jmig.2020.06.009&DestApp=DOI&SrcAppSID=USW2EC0A4ANOv50Iv8gX1hjIUlkOP&SrcJTitle=JOURNAL+OF+MINIMALLY+INVASIVE+GYNECOLOGY&DestDOIRegistrantName=Elsevier},
	doi = {10.1016/j.jmig.2020.06.009},
	language = {English},
	number = {7},
	urldate = {2024-09-01},
	journal = {Journal of Minimally Invasive Gynecology},
	author = {Harvey, Lara and Hendrick, Richard and Dillon, Neal and Blum, Evan and Branscombe, Lauren and Webster, Scott and Webster III, Robert J. and Anderson, Ted},
	month = dec,
	year = {2020},
	pages = {1631--1635},
}

@article{hendrick_hand-held_2015,
	title = {Hand-{Held} {Transendoscopic} {Robotic} {Manipulators}: {A} {Transurethral} {Laser} {Prostate} {Surgery} {Case} {Study}},
	volume = {34},
	issn = {0278-3649},
	shorttitle = {Hand-{Held} {Transendoscopic} {Robotic} {Manipulators}},
	url = {https://doi.org/10.1177/0278364915585397},
	doi = {10.1177/0278364915585397},
	number = {13},
	urldate = {2024-09-01},
	journal = {The International Journal of Robotics Research},
	author = {Hendrick, Richard J. and Mitchell, Christopher R. and Herrell III, S. Duke and Webster III, Robert J.},
	month = nov,
	year = {2015},
	pages = {1559--1572},
}

@article{dupont_design_2010,
	title = {Design and {Control} of {Concentric}-{Tube} {Robots}},
	volume = {26},
	issn = {1941-0468},
	doi = {10.1109/TRO.2009.2035740},
	number = {2},
	urldate = {2024-09-01},
	journal = {IEEE Transactions on Robotics},
	author = {Dupont, Pierre E. and Lock, Jesse and Itkowitz, Brandon and Butler, Evan},
	month = apr,
	year = {2010},
	pages = {209--225},
}

\appendices
\section{Linearized Residual Models}
\label{sec: jacobian_definitions}

We derive closed-form analytic Jacobians for all four residuals, enabling efficient Gauss-Newton linearization and direct reuse of curvature information within the SVN inference. All Jacobians are defined with respect to a left-multiplicative perturbation on $SE(3)$. As described previously, a perturbed pose is expressed as
\begin{equation}
    T(\Delta \xi) = \exp\!\left(\widehat{\Delta \xi}\right) T,
\end{equation}
where $\Delta \xi \in \mathbb{R}^6$ is a twist in the tangent space and $(\cdot)^\wedge$ denotes the mapping from $\mathbb{R}^6$ to the Lie algebra $\mathfrak{se}(3)$. For a generic residual $r(T)$, the Jacobian is defined as:
\begin{equation}
    J = \left.\frac{\partial r(T(\Delta\xi))}{\partial \Delta\xi}\right|_{\Delta\xi = 0}
\end{equation}
which maps an infinitesimal perturbation in the tangent space to a first-order change in the residual. The Jacobians were verified numerically using finite differences.

\subsection{Sparse Image Jacobian}
For a canonical 3D keypoint $X \in \mathbb{R}^3$ in the needle frame, the predicted point in the camera frame is $p(T) = TX = RX + t \in \mathbb{R}^3$ and its corresponding 2D prediction is $u(T) = \pi(p(T))$. Under a left-multiplicative perturbation and first-order approximation $\exp(\widehat{\Delta\xi}) \approx I + \widehat{\Delta\xi}$, the variation in $p$ is:
\begin{equation}
    \delta p \approx v - [p]_\times \omega
    \qquad \Rightarrow \qquad
    \frac{\partial p}{\partial \Delta\xi}
    =
    \begin{bmatrix}
        I & -[p]_\times
    \end{bmatrix}
\end{equation}
For a pinhole camera with intrinsics $(f_x, f_y, c_x, c_y)$, the projection 
Jacobian is:
\begin{equation}
    J_\pi(p) =
    \begin{bmatrix}
    \frac{f_x}{Z} & 0 & -\frac{f_x X}{Z^2} \\[4pt]
    0 & \frac{f_y}{Z} & -\frac{f_y Y}{Z^2}
    \end{bmatrix}
\end{equation}
By the chain rule, the full sparse image Jacobian is:
\begin{equation}
    J_s = J_\pi(p)
        \begin{bmatrix}
        I & -[p]_\times
        \end{bmatrix}
    \in \mathbb{R}^{2\times6}
\end{equation}
where $[\cdot]_\times$ denotes the skew-symmetric matrix. This expression 
applies independently to the tip and tail keypoints and is stacked to yield  $J_s \in \mathbb{R}^{4\times6}$ for the full residual $r_s(T) \in \mathbb{R}^4$.

\subsection{Dense Image Jacobian}
We wish to compute the Jacobian of the dense residual with respect to $\Delta\xi$.
For simplicity we consider a single point $x = x_i$ and write $C = C(T)$, $H = H(T)$, and $Q = Q_{plane}$.


Applying the product rule, we get:  
\begin{equation}
    \label{eq:xCx}
    \frac{\partial r_d}{\partial\Delta\xi}
    = x^T \frac{\partial C}{\partial\Delta\xi}\,x
      + 2x^T C \cancelto{0}{\frac{\partial x}{\partial\Delta\xi}}.
\end{equation}
The second term vanishes since the observed image points $x_i$ are fixed.
Expanding $C = H^{-T}QH^{-1}$ and applying the product rule:
\begin{equation}
    \frac{\partial C}{\partial\Delta\xi}
    = \frac{\partial H^{-T}Q}{\partial\Delta\xi} H^{-1}
      + H^{-T}Q \frac{\partial H^{-1}}{\partial\Delta\xi}
\end{equation}
Using the identities
$\partial Y^{-1}/\partial X = -Y^{-1}(\partial Y/\partial X)Y^{-1}$ and
$\partial Y^T/\partial X = (\partial Y/\partial X)^T$, the first term expands to:
\begin{align}
    \frac{\partial H^{-T}Q}{\partial\Delta\xi}
    &= \frac{\partial H^{-T}}{\partial\Delta\xi}Q
       + H^{-T}\cancelto{0}{\frac{\partial Q}{\partial\Delta\xi}} \notag \\
    &= H^{-T}\!\left(\frac{\partial H}{\partial\Delta\xi}\right)^{\!T} H^{-T}Q
\end{align}
Substituting back, the definition of $C$ appears in both terms:
\begin{align}
    \label{eq: HC-CH}
     \frac{\partial C}{\partial \Delta \xi} &= H^{-T}(\frac{\partial H}{\partial \Delta \xi})^T  \underbrace{H^{-T} Q H^{-1}}_{C} - \underbrace{H^{-T} QH^{-1}}_{C} \frac{\partial H}{\partial \Delta \xi} H^{-1}\\
     &= H^{-T}(\frac{\partial H}{\partial \Delta \xi})^T C - C \frac{\partial H}{\partial \Delta \xi} H^{-1}
\end{align}

From \ref{eq: H def}, we get $H(T') = P\,T'\,M$. Under the left-multiplicative update and first-order approximation of $\exp(\widehat{\Delta \xi}) = I+\widehat{\Delta \xi}$, we obtain:
\begin{gather}
H = P(I + \widehat{\Delta\xi})\,T_{\mathrm{est}}\,M
  = \underbrace{P\,T_{\mathrm{est}}\,M}_{H_0} + P\,\widehat{\Delta\xi}\,X, \\
\partial H = H - H_0 = P\,\widehat{\Delta\xi}\,X
\end{gather}
where $H_0 = H(T_{\mathrm{est}})$.

At this point, we wish to express the perturbation $\partial H$ in terms of the twist $\Delta \xi$. To do so requires factoring the twist out of the middle of the matrix product, so that each degree of freedom contributes linearly to $H$. We can express this differential by using the Lie algebra basis of $\mathfrak{se}(3)$:
\begin{equation}
    \widehat{\Delta \xi} = \sum^6_{j=1} \Delta \xi_j \hat{E_j}
\end{equation}
where each $\hat{E_j}$ is a generator matrix corresponding to one of the six motion directions (three rotational and three translational) \cite{blanco-claraco_tutorial_2022, eade_lie_nodate}. Thus, each element of $\Delta \xi$ becomes a scalar that can be extracted from the matrix products, at the cost of having to now sum across the 6 different generator matrices.




Substituting this into $\partial H$ and then evaluating the partial derivative with respect to each twist component is:
\begin{align}
\label{eq: HEj}
    \partial H &= \sum^6_{j=1} (P~\hat{E_j} ~X) \Delta \xi_j \\
    \frac{\partial H}{\partial \Delta \xi_j} &= P~\hat{E_j} ~X, \quad j = 1, \ldots, 6
\end{align}

One option to evaluate the final Jacobian is to substitute $\frac{\partial H}{\partial \Delta \xi_j}$ into \ref{eq: HC-CH} to get a closed-form solution: 
\begin{equation}
\label{eq: hc-ch0}
    \frac{\partial C}{\partial \Delta \xi_j} |_{0} = -H^{-T}(\frac{\partial H}{\partial \xi_j})^TC_0 - C_0 \frac{\partial H}{\partial \xi_j} H^{-1}
\end{equation}
where $C_0 = C(T_{est})$ represents the conic of the initial transformation estimate.

While sufficient, such a solution would still require explicit evaluation over the six generator matrices of $\mathfrak{se}(3)$, resulting in a Jacobian that is six times larger than necessary. To make the formulation more compact and computationally efficient, we apply the Kronecker identities: 
\begin{equation*}
    \mathrm{vec}(A^T) = K~\mathrm{vec}(A), \qquad \mathrm{vec}(ABC) = (C^T \otimes A)\mathrm{vec}(B)
\end{equation*}
where $A, B, C$ are matrices, $\mathrm{vec}(\cdot)$ is the column-wise vectorization operator, $K$ is a commutation matrix that transforms $\mathrm{vec}(A)$ into $\mathrm{vec}(A^ \mathrm{vec}T)$, and $\otimes$ denotes the Kronecker product. 

Returning to \ref{eq: HC-CH}, before substituting $\frac{\partial H}{\partial \Delta \xi_j}$, we can apply the Kronecker identity to express the derivative of $C$ as:
\begin{align*}
\label{eq: dc de}
    &\mathrm{vec}(\frac{\partial C}{\partial \Delta \xi}) =~\dots\\
    &= -(C^T \otimes H^{-T}) \mathrm{vec}(\frac{\partial H}{\partial \Delta \xi}^T) 
    -( H^{-T} \otimes C) \mathrm{vec}(\frac{\partial H}{\partial \Delta \xi})\\
    &= -(C^T \otimes H^{-T}) ~ K \mathrm{vec}(\frac{\partial H}{\partial \Delta \xi}) -( H^{-T} \otimes C) \mathrm{vec}(\frac{\partial H}{\partial \Delta \xi})\\
    & = -[(C^T \otimes H^{-T}) ~ K + ( H^{-T} \otimes C)] \mathrm{vec}(\frac{\partial H}{\partial \Delta \xi})
\end{align*}
Defining
\begin{equation}
D_H := -\bigl[(C^T\otimes H^{-T})\,K + (H^{-T}\otimes C)\bigr] \in \mathbb{R}^{9\times9}
\end{equation}
as the matrix $D_h$ represents the sensitivity of the conic $C$ with respect to infinitesimal perturbations in $H$. Next, we consider the term  $\mathrm{vec}(\frac{\partial H}{\partial \Delta \xi})$, which can vectorize the form with the basis matrices from  \ref{eq: HEj}:
\begin{equation}
    A := \bigl[\mathrm{vec}(P\hat{E}_1 X)\;\cdots\;\mathrm{vec}(P\hat{E}_6 X)\bigr]
    \in \mathbb{R}^{9\times6}
\end{equation}
where the matrix $A$ serves as the basis mapping from the twist perturbation to the differential in $H$, yielding: 
\begin{equation*}
    \mathrm{vec}(\frac{\partial C}{\partial \Delta \xi}) = D_H A
\end{equation*}

The Jacobian for a single backbone point follows from $r_{d,i} = \mathrm{vec}(C)^T\mathrm{vec}(x_i x_i^T)$:
\begin{equation}
    J_i = \mathrm{vec}(x_i x_i^T)^T D_H\,A \in \mathbb{R}^{1\times6}
\end{equation}

Stacking over all $m$ backbone points yields the full dense Jacobian:
\begin{equation}
    J_d = \frac{\partial r_d}{\partial\Delta\xi}\bigg|_{\Delta\xi=0}
    =
    \begin{bmatrix}
    \mathrm{vec}(x_1 x_1^T)^T \\
    \vdots \\
    \mathrm{vec}(x_m x_m^T)^T
    \end{bmatrix}
    D_H\,A
\end{equation}

\subsection{Robot Position Jacobian}
The robot position residual is $r_p(T) = TX_{grasp} - \hat{p}_{grasp}$, where $\hat{p}_{grasp}$ does not depend on $T$, so $J_p = \partial(TX_{grasp})/\partial\Delta\xi$.



Let $p_g = T_{\mathrm{est}}\,X_{grasp}$ denote the grasp point in the camera frame at the linearization point. Under the left-multiplicative update and first-order expansion, and using $a\times b = -[b]_\times a$:
\begin{align}
    p_g(T') &= \exp(\widehat{\Delta\xi})\,p_g \notag \\
             &\approx p_g + [\Delta\omega]_\times p_g + \Delta v
              = p_g - [p_g]_\times\,\Delta\omega + \Delta v
\end{align}
Taking the partial derivative:
\begin{equation}
J_p = \frac{\partial r_p}{\partial\Delta\xi}\bigg|_{\Delta\xi=0}
= \begin{bmatrix} -[p_g]_\times & I_3 \end{bmatrix} \in \mathbb{R}^{3\times6}
\end{equation}
where $p_g = T_{\mathrm{est}}\,X_{grasp}$.

\subsection{Robot Orientation Jacobian}
The robot orientation residual is $r_\perp(T) = \hat{z}_{grasper}^T\,\hat{t}_{grasp}(T)$, where $\hat{z}_{grasper} \in \mathbb{R}^3$ is the known grasper axis in the camera frame and $\hat{t}_{grasp}(T)$ is the unit needle tangent at $X_{grasp}$.

Let $t_0 \in \mathbb{R}^3$ denote the tangent direction at the grasp point in the needle frame. Under the pose $T=[R,t]$, the tangent direction in the camera frame is $t(T) = Rt_0$, and $\hat{t}_{grasp}(T) = t(T)/\|t(T)\|$. Since $t(T)$ depends only on $R$, the residual is independent of translation to first order.

Under the left-multiplicative perturbation, $R' \approx (I + [\omega]_\times)R$, so:
\begin{equation}
    t' = R' t_0 \approx t + \omega\times t
\end{equation}
giving,
\begin{equation}
    \delta t := t' - t \approx -[t]_\times\omega, \qquad
    \frac{\partial t}{\partial\Delta\xi}
    =
    \begin{bmatrix}
    0_{3\times3} & -[t]_\times
    \end{bmatrix}
\end{equation}

The differential of the normalization map $\hat{t} = t/\|t\|$ is $d\hat{t} = (1/\|t\|)(I - \hat{t}\hat{t}^T)\,dt$, so:
\begin{equation}
    \frac{\partial\hat{t}_{grasp}}{\partial\Delta\xi}
    = \frac{1}{\|t\|}\left(I - \hat{t}_{grasp}\hat{t}_{grasp}^T\right)
    \begin{bmatrix}
    0_{3\times3} & -[t]_\times
    \end{bmatrix}
\end{equation}


Applying the chain rule yields:
\begin{gather}
J_\perp
= \hat{z}_{grasper}^T
  \frac{1}{\|t\|}\left(I - \hat{t}_{grasp}\hat{t}_{grasp}^T\right)\\
  \begin{bmatrix}
  0_{3\times3} & -[t]_\times
  \end{bmatrix}
\in \mathbb{R}^{1\times6}.
\end{gather}
When $t_0$ is unit length, $\|t\| = 1$ and the expression simplifies accordingly.

\end{document}